\documentclass[11pt]{article}

\usepackage[final]{acl}

\usepackage{times}
\usepackage{latexsym}

\usepackage[T1]{fontenc}

\usepackage[utf8]{inputenc}

\usepackage{microtype}

\usepackage{inconsolata}

\usepackage{graphicx}

\usepackage{multirow}
\usepackage{colortbl}
\usepackage[table]{xcolor}
\usepackage{booktabs}
\usepackage{threeparttable}
\usepackage{graphicx}
\usepackage{amssymb}
\usepackage{wrapfig}
\usepackage{algorithm}
\usepackage{algpseudocode}
\usepackage{subfigure}
\usepackage[most]{tcolorbox}

\newtcbox{\highlight}{on line,
  colback=green!15!white,
  colframe=green!60!black,
  boxrule=0.5pt,
  arc=0pt,
  outer arc=0pt,
  boxsep=1pt,
  left=2pt,
  right=2pt,
  top=1pt,
  bottom=1pt}

\definecolor{teal}{rgb}{0.0,0.6,0.6}

\title{RACER: Retrieval-Augmented Contextual Rapid Speculative Decoding}

\author{
 \textbf{Zihong Zhang\textsuperscript{1}},
 \textbf{Zuchao Li\textsuperscript{1}\thanks{Corresponding author.}},
 \textbf{Lefei Zhang\textsuperscript{2}},
 \textbf{Ping Wang\textsuperscript{3}},
 \textbf{Hai Zhao\textsuperscript{4}},
\\
 \textsuperscript{1}School of Artificial Intelligence, Wuhan University, Wuhan, China \\
 \textsuperscript{2}School of Computer Science, Wuhan University, Wuhan, China \\
 \textsuperscript{3}School of Information Management, Wuhan University, Wuhan, China \\
 \textsuperscript{4}School of Computer Science, Shanghai Jiao Tong University, China
\\
 {\tt \{zhangzihong, zcli-charlie, zhanglefei, wangping\}@whu.edu.cn} \\
 {\tt zhaohai@cs.sjtu.edu.cn}
}

\begin{document}
\maketitle
\begin{abstract}
Autoregressive decoding in Large Language Models (LLMs) generates one token per step, causing high inference latency. Speculative decoding (SD) mitigates this through a guess-and-verify strategy, but existing training-free variants face trade-offs: retrieval-based drafts break when no exact match exists, while logits-based drafts lack structural guidance. We propose \textbf{RACER} (\textbf{R}etrieval-\textbf{A}ugmented \textbf{C}ont\textbf{e}xtual \textbf{R}apid Speculative Decoding), a lightweight and training-free method that integrates retrieved exact patterns with logit-driven future cues. This unification supplies both reliable anchors and flexible extrapolation, yielding richer speculative drafts. Experiments on Spec-Bench, HumanEval, and MGSM-ZH demonstrate that RACER consistently accelerates inference, achieving more than $2\times$ speedup over autoregressive decoding, and outperforms prior training-free methods, offering a scalable, plug-and-play solution for efficient LLM decoding. Our source code is available at \href{https://github.com/hkr04/RACER}{https://github.com/hkr04/RACER}.
\end{abstract}

\section{Introduction}

Large Language Models (LLMs) such as GPT~\citep{openai2025gptoss120bgptoss20bmodel}, LLaMA~\citep{dubey2024llama}, and Qwen~\citep{bai2023qwen} have achieved remarkable success across diverse natural language processing tasks. However, their autoregressive decoding paradigm, which generates one token per step, fundamentally limits inference efficiency. The sequential dependency causes inference latency to scale linearly with sequence length and model size, creating a key bottleneck for real-world deployment.

Speculative Decoding (SD) has emerged as a promising approach to address this challenge. By adopting a \textit{guess-and-verify} strategy, SD enables multiple tokens to be proposed and verified in parallel, achieving acceleration without sacrificing output quality. Existing methods fall into two categories. Model-based approaches rely on lightweight auxiliary models -- either separately trained~\citep{cai2024medusa, zhang2025hass, li2024eagle, li2024eagle2, li2025eagle3, shi2026scaling} or inherited from smaller variants of the same model family~\citep{leviathan2023sd, liu2025pearl} -- to generate draft tokens, at the cost of additional memory, training, and integration overhead. Model-free approaches, in contrast, construct draft tokens directly from signals available during inference. Among model-free approaches, most are retrieval-based, leveraging exact token sequence matches from static corpora or dynamically generated contexts~\citep{saxena2023pld, he2023rest}. Recent work further exploits the predictive power of last logit~\citep{liu2025logitspec}, or recycles candidate logits~\citep{luo-etal-2025-token-recycling}, showing that LLMs inherently encode richer cues for near-future tokens than previously assumed.

Despite these advances in model-free methods, two key limitations remain. First, retrieval-based methods depend on exact token matching, which breaks down when no continuation can be directly aligned. Second, logits-based methods are restricted to last-step or self-drafted candidates and lack external structural guidance, making it difficult to extrapolate toward more suitable tokens. As a result, their predictions tend to be narrow in scope and suboptimal in quality.

To address these limitations, we propose \textbf{RACER} (\textbf{R}etrieval-\textbf{A}ugmented \textbf{C}ont\textbf{e}xtual \textbf{R}apid Speculative Decoding), a \textbf{plug-and-play, training-free} method that unifies the strengths of both paradigms. Retrieval provides \textbf{seen information} through exact pattern matches, offering structural guidance, while logits supply \textbf{unseen information}, enabling extrapolation beyond strict matches. By augmenting logit predictions with retrieval signals, RACER generates richer and more accurate speculative drafts. In this way, retrieval functions not as an independent generator but as structural guidance that empowers logits to hypothesize plausible continuations beyond their immediate horizon.

We conduct comprehensive experiments across general benchmark Spec-Bench~\citep{xia-etal-2024-specbench}, code generation benchmark HumanEval~\citep{chen-etal-2021-humaneval} and Chinese math reasoning benchmark MGSM-ZH~\citep{shi2022mgsm}. Overall, we demonstrate that RACER provides a lightweight, training-free SD solution that effectively leverages complementary seen and unseen information via a unified cache-like module, delivering consistent inference acceleration with stable memory usage.

\section{Background}

\paragraph{Speculative Decoding}
Speculative Decoding (SD) typically proceeds in two phases: a \emph{drafting phase} and a \emph{verification phase}. 

Given a prefix $\mathbf{x}$, a lightweight \textbf{draft model} $M_q$ generates $\gamma$ candidate tokens 
$\tilde{x}_1, \ldots, \tilde{x}_\gamma$. 
These tokens, together with the prefix, are then passed to the \textbf{target model} $M_p$, which produces logits 
$p_1, \ldots, p_{\gamma}$. 
Each draft token $\tilde{x}_i$ is verified by comparing its probability under $M_p$ with that under $M_q$~\citep{leviathan2023sd, chen2023sd}:  

\begin{equation}
\alpha_i = 
\begin{cases}
1 & \text{if } p_i[\tilde{x}_i] \ge q_i[\tilde{x}_i], \\
\dfrac{p_i[\tilde{x}_i]}{q_i[\tilde{x}_i]} & \text{otherwise.}
\end{cases}
\end{equation}

If $\tilde{x}_i$ is accepted (with probability $\alpha_i$), it is appended to the sequence; otherwise, $\tilde{x}_i, \ldots, \tilde{x}_\gamma$ are discarded, and the SD step terminates early.
The next iteration then resumes from the last accepted prefix, using the last accepted logits $p_{i-1}$ of the target model to resample $x_i$ as the new continuation token. 
This guarantees that every iteration produces at least one token, while leveraging accepted drafts whenever possible to accelerate generation.

Regardless of whether greedy or nucleus sampling is employed, validation always leverages the logits from the target model.  
In expectation, a single SD step can advance by up to $\gamma+1$ tokens, significantly reducing the number of target model invocations compared to standard autoregressive decoding.

\paragraph{Retrieval-based Methods}
Retrieval-based SD methods bypass the draft model $M_q$ and instead rely on pattern matching within the token sequence. 
PLD~\citep{saxena2023pld}, for example, stores past $n$-grams together with their succeeding $m$-grams as predictions. 
This method is simple and effective in pattern-repeating scenarios such as code generation, 
but can only propose a single continuation at a time. 
Moreover, because pattern matches are sparse and fail to capture the full diversity of target model outputs, 
PLD is constrained to specific domains and cannot generalize broadly.

\paragraph{Tree Attention}
The standard \textit{guess-and-verify} scheme assumes a linear draft sequence. 
\emph{Tree attention}~\citep{miao2024specinfer, cai2024medusa} generalizes this by allowing the draft model to propose a branching tree of candidates. 
During verification, the target model processes all nodes in parallel, 
with position encodings set by depth and attention masks restricting each node to its ancestors:  
\begin{equation}
\begin{aligned}
& \mathrm{pos}[i] = \mathrm{pos}[\mathrm{parent}(i)] + 1, \\
& \mathrm{mask}[i,j] = \mathbb{1}[\, j=i \ \text{or}\ j \in \text{ancestor}(i) \,].
\end{aligned}
\end{equation}
This transforms SD into a branching search process, 
enabling higher parallelism and more effective utilization of the target model when multiple plausible continuations exist.

For model-based methods, Medusa~\citep{cai2024medusa} attaches multiple additional language model (LM) heads to the top layer, each predicting draft tokens for different tree depths. 
EAGLE-3~\citep{li2025eagle3} further integrates low-, mid-, and high-level features of the target model, with its core structured as a Transformer decoder layer. 
For model-free methods, REST~\citep{he2023rest} constructs a suffix array to identify the longest suffix match and expands the matched continuation into a trie, 
while SAM Decoding~\citep{hu-etal-2025-sam} employs both dynamic and static suffix automata to capture contextual as well as pre-built suffix patterns, providing more flexible retrieval-guided expansions.

\section{Methodology}

In this section, we introduce our approach in three parts: 
\emph{Logits Tree}, \emph{Retrieval Tree with LRU (Least Recently Used) eviction}, and their integration strategy.

\begin{figure*}[htbp]
    \centering
    \includegraphics[width=1\linewidth]{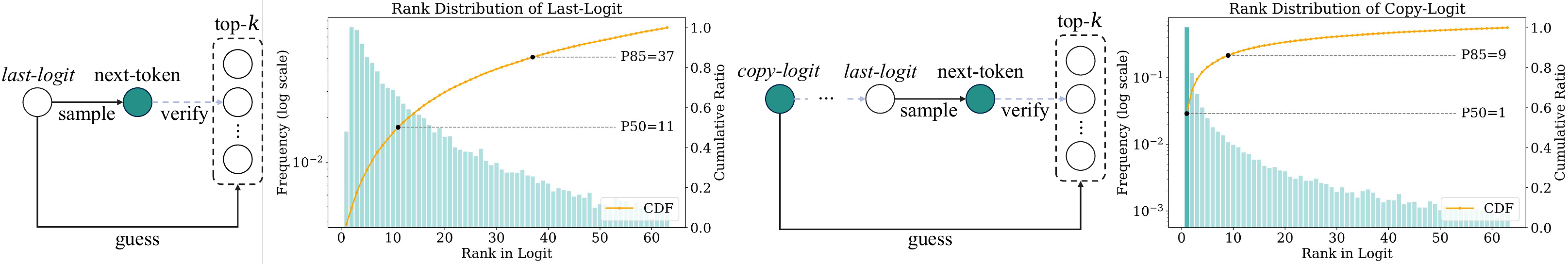}
    \caption{The \textit{last-logit} node (white) produces both the next-token sample and the draft tokens immediately after it. The \textit{copy-logit} node (green) marks the same token ID as the next-token, whose logit is reused to approximate the next-token's logits when generating subsequent draft tokens.}
    \label{fig:first_layer}
\end{figure*}

\subsection{Logits Tree}
\label{sec:logits_tree}

\paragraph{First Step beyond Next-Token}
We examine two \textit{logit-reuse} strategies when extending the tree beyond the next token. Let the verified prefix be $x_{<t}$, and let the target model produce logits $\mathbf{z}_t = f(x_{<t})$. The next token is sampled as $x_t \sim \text{Softmax}(\mathbf{z}_t)$ without computing $\mathbf{z}_{t+1}$.
To approximate the unknown future logits, it's natural to construct a surrogate distribution $\tilde{\mathbf{z}}_{t+1}$ and expand its top-$k$ tokens as candidates for $x_{t+1} \sim \text{Softmax}(\mathbf{z}_{t+1})$:
$$
\tilde{x}_{t+1}^{(k)} = \text{TopK}_k\big(\text{Softmax}(\tilde{\mathbf{z}}_{t+1})\big).
$$
We define two strategies: \textit{last-logit} with $\tilde{\mathbf{z}}_{t+1} = \mathbf{z}_t$, and \textit{copy-logit} with $\tilde{\mathbf{z}}_{t+1} = \mathbf{z}_{i+1}$ where $i < t$ is the nearest index such that $x_i = x_t$.

The \textit{last-logit} strategy reuses the logit distribution from which the next-token was sampled to expand all of its children candidates, assuming local smoothness in the token space.
The \textit{copy-logit} strategy instead reuses the logit from the most recent occurrence of the same token, assuming that identical vocabulary tokens tend to preserve similar semantic tendencies when appearing in comparable contexts.

To evaluate these strategies, we conducted experiments on Spec-Bench using Vicuna-7B, OpenPangu-7B and Qwen3-1.7B under greedy decoding. 
For each speculative step, we selected the top-63 tokens from the corresponding logits to form the first layer, 
and measured their effectiveness by the mean accepted tokens (MAT) and the distribution of accepted ranks (from 1 to 63). 
Since all models exhibited similar trends, we report only the results of Vicuna-7B here. Additional results and discussions can be found in Appendix~\ref{sec:explanation_logits_tree}.

\begin{figure}[htbp]
    \centering
    \includegraphics[width=1\columnwidth]{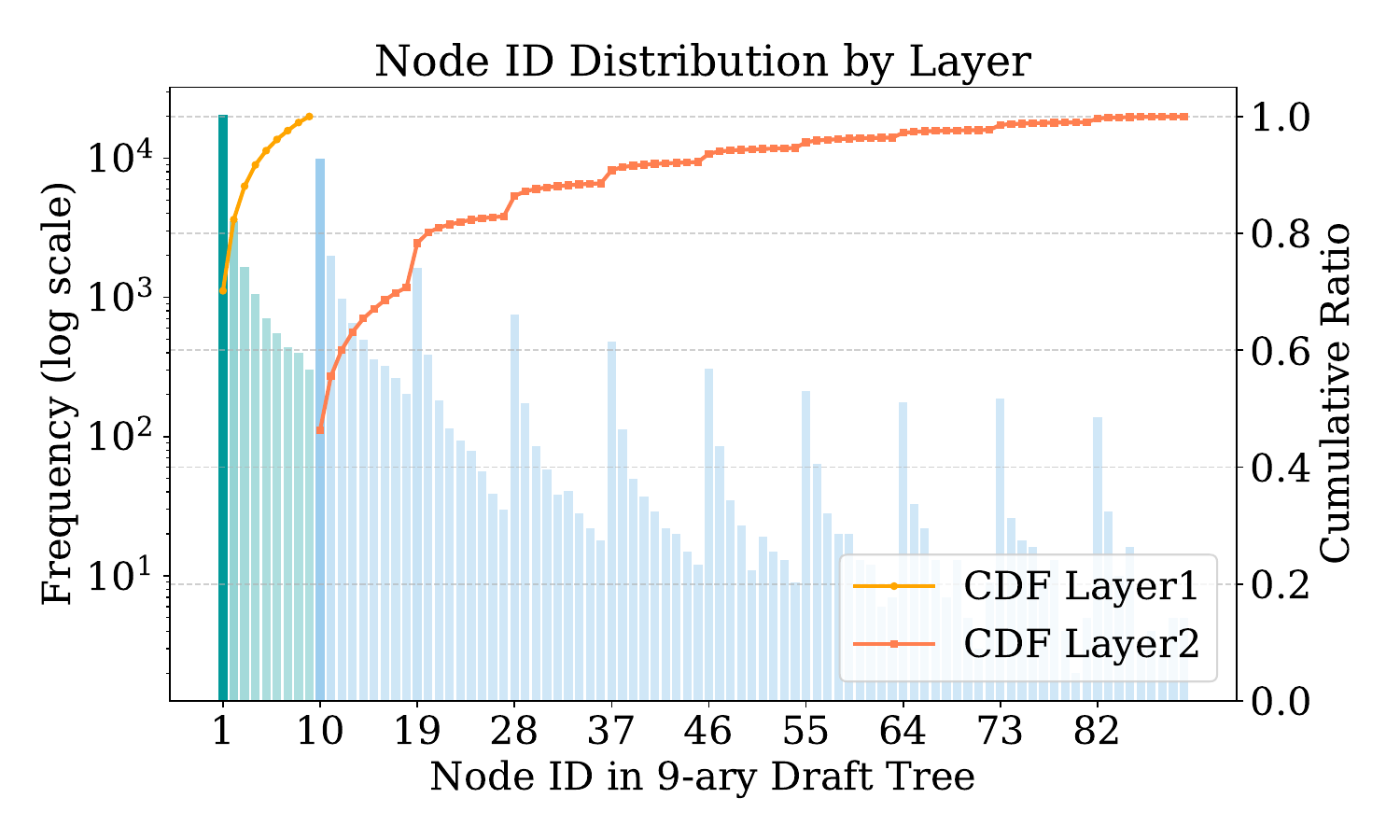}
    \caption{
        Analogy experiments with a fixed $9$-ary draft tree of height $3$. 
    }
    \label{fig:kary_tree}
\end{figure}

The MAT values of \textit{last-logit} and \textit{copy-logit} are 1.57 and 1.87, respectively, 
indicating that \textit{copy-logit} achieves higher acceptance rate. 
As shown in Figure~\ref{fig:first_layer}, the \textit{copy-logit} strategy also exhibits a pronounced heavy-tail property: 
its accepted tokens concentrate strongly at the top ranks, with rank-1 alone accounting for more than $50\%$ of accepted cases. 
For \textit{copy-logit}, the 50th and 85th percentile accepted ranks are 1 and 9, compared with 11 and 37 for \textit{last-logit}.
These results demonstrate that \textit{copy-logit} provides a sharper and more reliable distribution for speculative expansion, 
and we therefore adopt it as the basic expansion strategy of Logits Tree.

\begin{figure*}[htbp]
    \centering
    \includegraphics[width=1\linewidth]{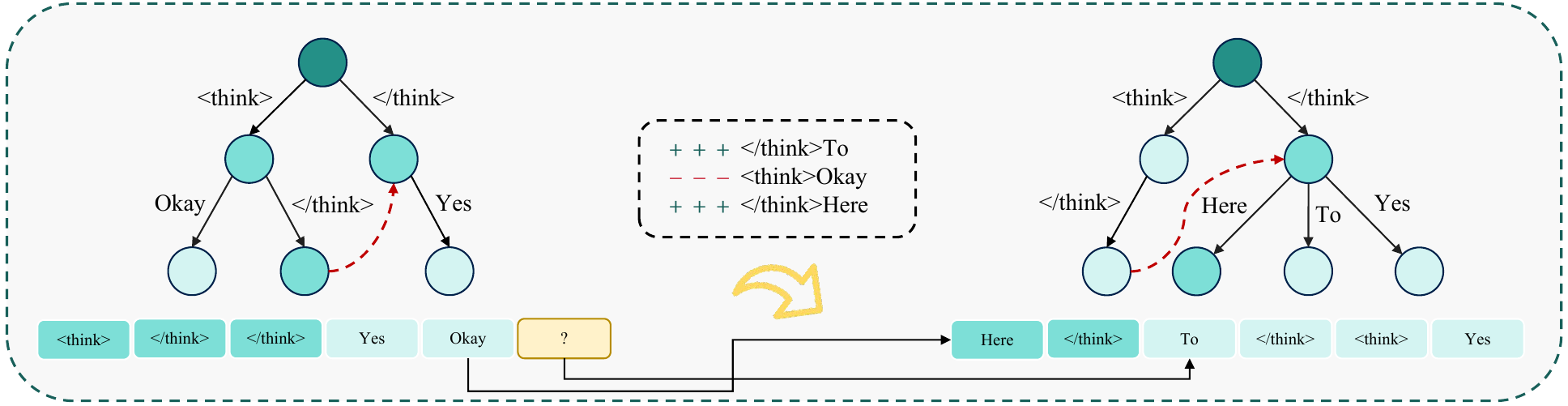}
    \caption{Illustration of the LRU-based eviction strategy in RACER's retrieval automaton. Solid black edges denote standard trie transitions, and dashed red edges denote failure links.
    Yellow nodes represent unallocated states.
    Green nodes indicate allocated states, where darker color corresponds to more recent usage.
    The example demonstrates how inserting the 2-gram \texttt{[</think>, To]} creates a new node \texttt{To}, and how LRU leaf node \texttt{Okay} is evicted and replaced with \texttt{Here} when the capacity is reached with the new 2-gram \texttt{[</think>, Here]}.}
    \label{fig:automaton_lru}
\end{figure*}

\paragraph{$k$-ary Analogy and Pruning}
Motivated by the heavy-tail property observed in Figure~\ref{fig:first_layer}, 
we next extend the expansion recursively to deeper layers. 
Two principles guide this design: 
(i) the 85th percentile rank is around 9, indicating that useful candidates concentrate in the head of the distribution; 
and (ii) because SD proceeds prefix-wise, the breadth of any child node should not exceed that of the root expansion.

To further examine the implications of the above principles and to explore how to prune the draft tree effectively, 
we conducted analogy experiments with a fixed $9$-ary draft tree of height $3$. 
This tree contains $1 + 9 + 9^2 = 91$ nodes, indexed from $0$ to $90$.
Node 0 is the root corresponding to the next-token position; nodes 1-9 form the first layer; and nodes 10-90 form the second layer. Under this level-order indexing scheme, each node $i > 0$ has its parent given by $\mathrm{parent}(i) = \lfloor\frac{i - 1}{k}\rfloor$.

As illustrated in Figure~\ref{fig:kary_tree}, the second-layer children of node~1 exhibit a cumulative trend resembling that of the first layer, albeit slightly slower. For children of parents with larger IDs, the growth further slows and the total volume is nearly halved. The distribution remains front-loaded. Compared with \textit{copy-logit}, the MAT increases from 1.87 to 2.34.

To capture this behavior, we define the breadth allocation in the Logits Tree for child nodes as
\begin{equation}
\label{eq:breadth}
\begin{aligned}
b_{\mathrm{child}(i, j)}
&= \max\!\left(
    1,
    \left\lfloor \tfrac{b_i}{2^{j + [i \neq 0]}} \right\rfloor
\right), \\
&\qquad j = 0, 1, \ldots, b_i - 1.
\end{aligned}
\end{equation}

where $b_i$ is the breadth of the parent node and $j$ is the child index. 
Specifically, nodes at the first layer start with the maximum breadth, while deeper layers inherit half of their parent's breadth.
This design ensures that the upper part of the Logits Tree expands more aggressively, while deeper layers are progressively pruned.
Given a specific draft capacity, the Logits Tree then expands in a breadth-first manner according to this allocation rule. Figure~\ref{fig:logits_tree} in the Appendix illustrates this process using a 4-ary example, showing: (i) the original $k$-ary indexing, and (ii) the pruned Logits Tree structure.

\subsection{Retrieval Tree with LRU Eviction}
\label{sec:retrieval}

Approaches such as SAM Decoding~\citep{hu-etal-2025-sam} and LogitSpec~\citep{liu2025logitspec} indicate that explicit retrieval drafts can complement the logits and improve acceptance rates. 
Motivated by these findings, we aim to design an efficient retrieval structure that exploits repeated patterns in the context.
Classical indexing structures such as suffix arrays~\citep{manber1993suffix} or suffix automata~\citep{blumer1984sam} provide efficient substring matching, 
but they grow proportionally with the context length and lack a natural mechanism to discard obsolete states. 
This is undesirable in language modeling, where the distribution of substrings follows a Zipf-like long-tail law, 
implying that many substrings have little utility and can be safely evicted.
To balance efficiency and adaptivity, we propose to use an Aho--Corasick (AC) automaton~\citep{aho1975efficient} to maintain an $n$-gram-based Retrieval Tree. 
Unlike a plain $n$-gram trie, the AC automaton supports failure links that facilitate fast state transitions and naturally enrich draft diversity. See Appendix~\ref{sec:ac_automata} for a brief introduction to the AC automaton.

\paragraph{Transition Rule} As shown in Figure~\ref{fig:automaton_lru}, we incorporate an \textbf{LRU-style (Least Recently Used) eviction mechanism} into the AC automaton so that infrequent $n$-gram patterns are pruned while new ones from the incoming context are continually incorporated. Each time a state is visited -- either through a valid transition or via a failure-link fallback -- that state is marked as ``touched''.
Importantly, when backtracking with failure links, all of its ancestor (prefix) states are also necessarily touched. This ensures that prefix nodes always remain equal or more recent than their descendants.
For example, when the token \texttt{Yes} follows \texttt{[<think>, </think>]} and there is no direct transition, the automaton backtracks along its failure link to state \texttt{[</think>]} and then transitions to \texttt{[</think>, Yes]}; all states along this fallback and transition path are touched.

\begin{figure*}[htbp]
    \centering
    \includegraphics[width=1\linewidth]{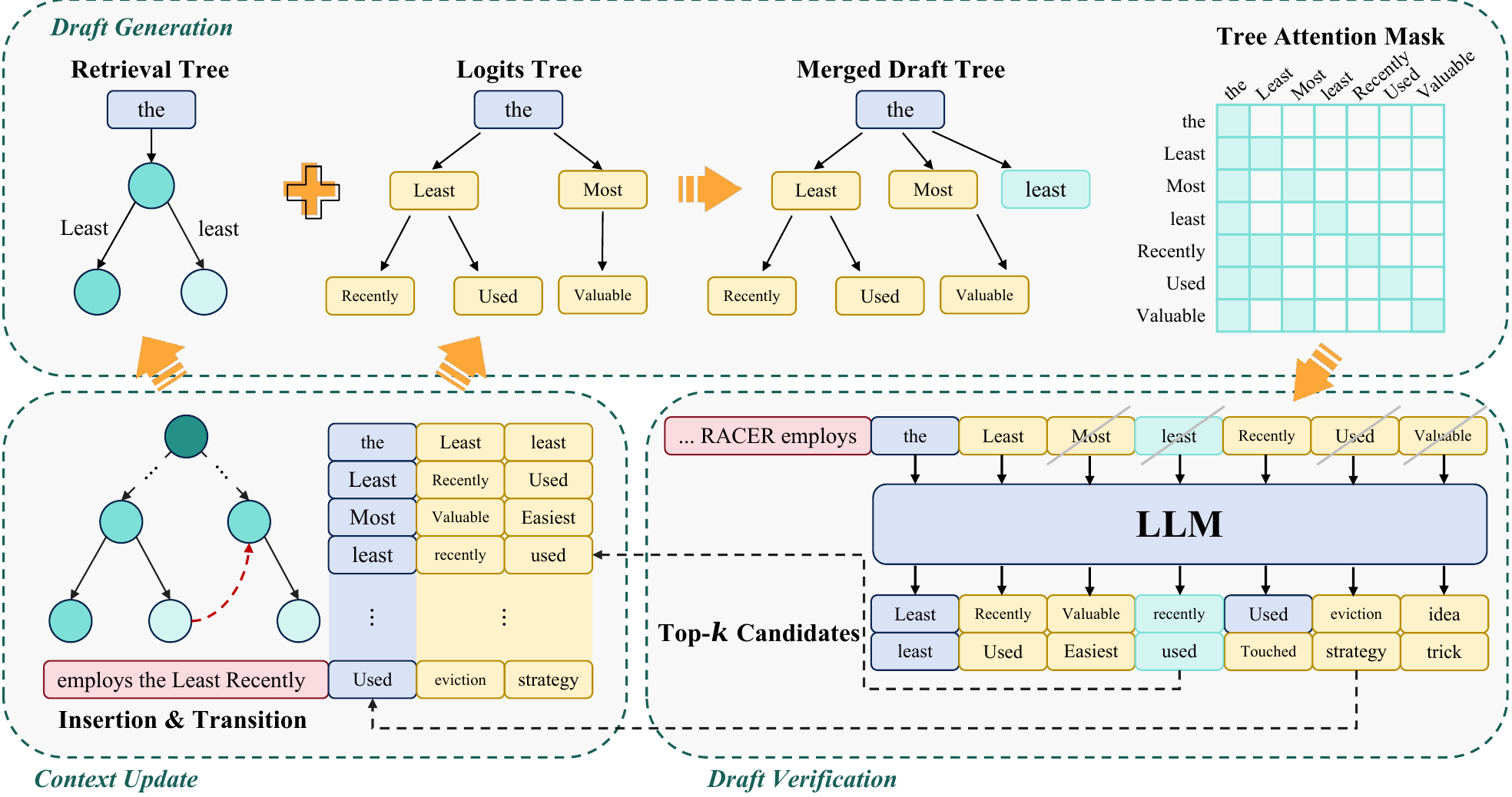}
    \caption{Overview of RACER. At each decoding step, the AC automaton accepts the next token and identifies border nodes with depth $\geq 2$, from which the globally most frequent children are selected as retrieval candidates. If retrieval nodes do not fill the draft capacity $C$, the remaining slots are assigned to Logits Tree expansion (Eq.~\ref{eq:breadth}). Verified $n$-grams are inserted into the automaton, while the logits adjacency matrix is refreshed with the newly generated logits.}
    \label{fig:framework}
\end{figure*}

\paragraph{Update and Eviction Rule}
During decoding, newly observed $n$-grams are incrementally inserted into the automaton.  
Insertion follows the transition path whenever possible; if a transition does not exist, a new node is allocated either from an empty slot or by reusing an LRU leaf node.  
When the automaton reaches its predefined capacity, the LRU \emph{leaf} node is evicted (e.g., the leaf \texttt{Okay} in Figure~\ref{fig:automaton_lru}).  
All nodes are managed using a hash table and a doubly linked list, enabling $\mathcal{O}(1)$ updates and eviction.
Failure links are updated lazily: they are rebuilt only once at the end of the prefilling phase.  
Before the rebuild, the newly added portion of the structure temporarily behaves as a standard trie without failure links.

\paragraph{Expansion Rule}
We consider all match states (borders) whose matched depth is at least $2$.  
For each border, we take the sub-trie rooted at that state, collect all outgoing continuations, pool them across borders, and select the globally most frequent top-$k$ continuation states to expand the Retrieval Tree. Detailed examples of the expansion procedure based on the frequencies are presented in Appendix~\ref{sec:case_study_expansion}.

\subsection{Integration Strategy}

Given a fixed speculative capacity $C$, retrieval-based candidates are first generated according to the expansion rule in Section~\ref{sec:retrieval}, 
while the remaining capacity is allocated to the Logits Tree via breadth-first expansion (Eq.~\ref{eq:breadth}). 
Since retrieval candidates are sparse but structurally reliable, 
we retain only the most confident ones and leave the remaining budget 
for logits-based exploration through the top-$k$ adjacency matrix. 

Importantly, retrieval not only complements the current speculation, 
but also provides stronger cues for upcoming tokens than logits alone. 
Because retrieval captures repeated patterns in closer contexts, 
it guides the logits distribution toward sharper predictions 
and mitigates error accumulation in speculative expansions. 

The two candidate sets are finally merged into a unified draft tree through a trie-based union, 
and verified by the target model under the \textit{guess-and-verify} scheme. 
This hybrid design enables RACER to exploit both \emph{seen information} from retrieval 
and \emph{unseen speculation} from logits, achieving higher acceptance rates with controlled memory usage. The overall workflow is presented in Figure~\ref{fig:framework}; further implementation details can be found in Section~\ref{sec:impl_detail}.
\section{Experiments}
\label{sec:experiments}

\begin{table*}[htbp]
\centering
\caption{Results on Spec-Bench, HumanEval, and MGSM-ZH 
with \textbf{Vicuna} and \textbf{LLaMA3.1} target models.
Reported in MAT and speedup ratio.
Best results are in \textbf{bold}, suboptimal results are \underline{underlined}.
}
\small
\label{tab:mix_main_instruct}
\begin{threeparttable}
\renewcommand{\arraystretch}{1.2}
\setlength{\tabcolsep}{5pt}
\begin{tabular}{l l cc cc cc cc}
\toprule
\multirow{2}{*}{\textbf{Model}} & \multirow{2}{*}{\textbf{Method}}
& \multicolumn{2}{c}{\textbf{Spec-Bench}} 
& \multicolumn{2}{c}{\textbf{HumanEval}} 
& \multicolumn{2}{c}{\textbf{MGSM-ZH}} 
& \multicolumn{2}{c}{\textbf{Average}} \\
\cmidrule(lr){3-4} \cmidrule(lr){5-6} \cmidrule(lr){7-8} \cmidrule(lr){9-10}
& & MAT & Speedup & MAT & Speedup & MAT & Speedup & MAT & Speedup \\
\midrule
\multirow{5}{*}{Vicuna 7B}
 & PLD       & 1.71 & 1.50 & 1.58 & 1.40 & 2.57 & 2.27 & 1.95 & 1.87 \\
 & REST      & 1.82 & 1.45 & 2.06 & 1.71 & 1.29 & 1.06 & 1.72 & 1.41 \\
 & LogitSpec & 2.34 & 1.77 & 2.22 & 1.66 & \underline{3.55} & \underline{2.67} & 2.70 & 2.03 \\
 & TR        & \underline{2.76} & \underline{2.06} & \underline{2.83} & \underline{2.17} & 3.00 & 2.30 & \underline{2.86} & \underline{2.18} \\
\rowcolor{teal!20}\cellcolor{white}
 & \textbf{RACER} & \textbf{3.00} & \textbf{2.21} & \textbf{3.11} & \textbf{2.29} & \textbf{3.71} & \textbf{2.77} & \textbf{3.27} & \textbf{2.42} \\
\midrule
\multirow{5}{*}{Vicuna 13B}
 & PLD       & 1.65 & 1.41 & 1.59 & 1.43 & 2.45 & 2.11 & 1.90 & 1.65 \\
 & REST      & 1.82 & 1.44 & 2.07 & 1.71 & 1.31 & 1.07 & 1.73 & 1.41 \\
 & LogitSpec & 2.32 & 1.73 & 2.23 & 1.77 & \underline{3.44} & \underline{2.72} & 2.66 & 2.07 \\
 & TR        & \underline{2.79} & \underline{1.99} & \underline{2.83} & \underline{2.08} & 3.05 & 2.22 & \underline{2.89} & \underline{2.10} \\
\rowcolor{teal!20}\cellcolor{white}
 & \textbf{RACER} & \textbf{2.95} & \textbf{2.25} & \textbf{3.09} & \textbf{2.42} & \textbf{3.64} & \textbf{2.83} & \textbf{3.23} & \textbf{2.50} \\
\midrule
\multirow{5}{*}{Vicuna 33B}
 & PLD       & 1.33 & 1.03 & 1.64 & 1.48 & 2.18 & 1.97 & 1.72 & 1.49 \\
 & REST      & 1.81 & 1.54 & 1.98 & 1.72 & 1.32 & 1.17 & 1.70 & 1.48 \\
 & LogitSpec & 2.32 & 1.73 & 2.35 & 1.92 & \underline{2.96} & \underline{2.44} & 2.54 & \underline{2.03} \\
 & TR        & \underline{2.63} & \underline{1.83} & \underline{2.79} & \underline{2.05} & 2.83 & 2.10 & \underline{2.75} & 1.99 \\
\rowcolor{teal!20}\cellcolor{white}
 & \textbf{RACER} & \textbf{2.74} & \textbf{2.20} & \textbf{3.16} & \textbf{2.58} & \textbf{3.36} & \textbf{2.77} & \textbf{3.09} & \textbf{2.52} \\
\midrule
\multirow{2}{*}{LLaMA3.1 8B}
 & EAGLE-3 & \textbf{3.76} & \textbf{2.51} & \textbf{4.41} & \textbf{3.06} & \underline{1.54} & \underline{1.06} & \textbf{3.24} & \underline{2.21} \\
 & \cellcolor{teal!20}\textbf{RACER} & \cellcolor{teal!20}\underline{2.82} & \cellcolor{teal!20}\underline{2.41} & \cellcolor{teal!20}\underline{3.22} & \cellcolor{teal!20}\underline{2.87} & \cellcolor{teal!20}\textbf{3.33} & \cellcolor{teal!20}\textbf{2.89} & \cellcolor{teal!20}\underline{3.12} & \cellcolor{teal!20}\textbf{2.72} \\
\bottomrule
\end{tabular}
\end{threeparttable}
\end{table*}

\subsection{Experimental Setup}

Following prior work~\citep{luo-etal-2025-token-recycling}, we focus on greedy decoding with batch size $1$ and maximum output length $1024$. We report the following metrics: \textbf{Mean Accepted Tokens (MAT)}: the average number of tokens confirmed in a single speculative decoding step; \textbf{Speedup Ratio}: relative performance compared with HuggingFace's implementation of autoregressive decoding.

For all experiments, we use the following default hyperparameters unless otherwise specified. 
For the Logits Tree, the maximum breadth is set to $8$. 
For the Retrieval Tree, we maintain up to $10{,}000$ nodes with an $n$-gram length of $10$. 
The draft size of each decoding step is 64 as suggested in Medusa~\citep{cai2024medusa}.
These values are chosen based on preliminary analyses, 
and we further demonstrate in Appendix~\ref{sec:parameter_robustness} that our method is robust with respect to these hyperparameters. Further details can be found in Appendix~\ref{sec:appendix_setup}.

\paragraph{Target Models and Datasets}
We utilize instruct model Vicuna~\citep{vicuna2023} at three scales: 7B, 13B, and 33B, 
where the 7B and 13B models use version~1.5 and the 33B model uses version~1.3. In addition, we include LLaMA3.1-8B.
We further incorporate reasoning models: OpenPangu~\citep{chen2025pangu} at 7B and Qwen3~\citep{yang2025qwen3} at the corresponding 8B, 14B, and 32B scales.
We evaluate on three benchmarks: 
\textbf{Spec-Bench}~\citep{xia-etal-2024-specbench}, 
\textbf{HumanEval}~\citep{chen-etal-2021-humaneval}, 
and \textbf{MGSM-ZH}~\citep{cobbe2021gsm8k, shi2022mgsm}. 
Spec-Bench covers diverse scenarios including Multi-turn Conversation (MT), Translation (Trans), Summarization (Sum), Question Answering (QA), 
Mathematical Reasoning (Math), and Retrieval-Augmented Generation (RAG). 
HumanEval is a widely used benchmark for code generation. 
MGSM-ZH is the Chinese counterpart of GSM8K. Following the multilingual evaluation setup used in PEARL~\citep{liu2025pearl}, we adopt MGSM-ZH to assess non-English mathematical reasoning in a language where OpenPangu and Qwen3 exhibit stable performance, while avoiding languages for which Vicuna generates unstable outputs.
Together, these benchmarks cover general-purpose, domain-specific, and cross-lingual reasoning tasks.

\begin{table*}[htbp]
\centering
\caption{
Results on Spec-Bench, HumanEval, and MGSM-ZH 
with \textbf{OpenPangu} and \textbf{Qwen3} target models.
Reported in MAT and speedup ratio.
Best results are in \textbf{bold}, suboptimal results are \underline{underlined}.
}
\small
\label{tab:mix_main_reasoning}
\begin{threeparttable}
\renewcommand{\arraystretch}{1.2}
\setlength{\tabcolsep}{5pt}
\begin{tabular}{l l cc cc cc cc}
\toprule
\multirow{2}{*}{\textbf{Model}} & \multirow{2}{*}{\textbf{Method}}
& \multicolumn{2}{c}{\textbf{Spec-Bench}} 
& \multicolumn{2}{c}{\textbf{HumanEval}} 
& \multicolumn{2}{c}{\textbf{MGSM-ZH}} 
& \multicolumn{2}{c}{\textbf{Average}} \\
\cmidrule(lr){3-4} \cmidrule(lr){5-6} \cmidrule(lr){7-8} \cmidrule(lr){9-10}
& & MAT & Speedup & MAT & Speedup & MAT & Speedup & MAT & Speedup \\
\midrule
\multirow{2}{*}{OpenPangu 7B}
 & PLD       & \underline{1.48} & \underline{1.37} & \underline{1.44} & \underline{1.27} & \underline{1.53} & \underline{1.41} & \underline{1.47} & \underline{1.35} \\
 & \cellcolor{teal!20}\textbf{RACER} & \cellcolor{teal!20}\textbf{2.47} & \cellcolor{teal!20}\textbf{1.99} & \cellcolor{teal!20}\textbf{2.65} & \cellcolor{teal!20}\textbf{2.12} & \cellcolor{teal!20}\textbf{2.77} & \cellcolor{teal!20}\textbf{2.26} & \cellcolor{teal!20}\textbf{2.63} & \cellcolor{teal!20}\textbf{2.12}\\
\midrule
\multirow{3}{*}{Qwen3 8B}
 & PLD       & 1.52 & 1.35 & 1.52 & 1.41 & \underline{1.69} & \underline{1.52} & 1.58 & 1.43 \\
 & EAGLE-3$^\dagger$ & \textbf{3.46} & \textbf{2.14} & \textbf{3.84} & \textbf{2.44} & 1.41 & 0.86 & \textbf{2.90} & \underline{1.81} \\
\rowcolor{teal!20}\cellcolor{white}
 & \textbf{RACER} & \underline{2.73} & \underline{2.13} & \underline{2.79} & \underline{2.24} & \textbf{2.95} & \textbf{2.26} & \underline{2.82} & \textbf{2.21} \\
\midrule
\multirow{3}{*}{Qwen3 14B}
 & PLD       & 1.45 & 1.34 & 1.43 & 1.27 & \underline{1.59} & \underline{1.49} & 1.49 & 1.37 \\
 & EAGLE-3$^\dagger$   & \textbf{2.72} & \underline{1.87} & \textbf{3.03} & \underline{2.05} & 1.56 & 1.12 & \underline{2.44} & \underline{1.68} \\
\rowcolor{teal!20}\cellcolor{white}
 & \textbf{RACER} & \underline{2.67} & \textbf{2.23} & \underline{2.77} & \textbf{2.29} & \textbf{2.88} & \textbf{2.30} & \textbf{2.77} & \textbf{2.27} \\
\midrule
\multirow{3}{*}{Qwen3 32B}
 & PLD       & 1.45 & 1.34 & 1.34 & 1.23 & 1.56 & \underline{1.46} & 1.45 & 1.34 \\
 & EAGLE-3$^\dagger$  & \textbf{2.88} & \underline{2.12} & \textbf{2.97} & \textbf{2.17} & \underline{1.60} & 1.18 & \underline{2.48} & \underline{1.82} \\
\rowcolor{teal!20}\cellcolor{white}
 & \textbf{RACER} & \underline{2.66} & \textbf{2.17} & \underline{2.55} & \underline{2.08} & \textbf{2.78} & \textbf{2.28} & \textbf{2.66} & \textbf{2.18} \\
\bottomrule
\end{tabular}
\begin{tablenotes}
\footnotesize
\item[$\dagger$] EAGLE-3 models and weights from AngelSlim's re-implementation.
\end{tablenotes}
\end{threeparttable}
\end{table*}

\paragraph{Baselines}
We compare RACER against two retrieval-based methods (PLD and REST), 
two logits-involved methods (Token Recycling and LogitSpec), 
and the state-of-the-art model-based method EAGLE-3. 
\textbf{PLD}~\citep{saxena2023pld} stores past $n$-grams as sequential keys 
and their succeeding $m$-grams as predicted values.  
\textbf{REST}~\citep{he2023rest} builds a suffix array over the training set 
to locate the longest suffix match, then expands the matched continuation into a trie.  
\textbf{Token Recycling (TR)}~\citep{luo-etal-2025-token-recycling} maintains a top-$k$ adjacency matrix 
from token logits and extends it into a draft tree using a predefined template.  
\textbf{LogitSpec}~\citep{liu2025logitspec} speculates the first draft token from the top-$k$ candidates 
of the last-step logits, then augments expansion with retrieval tokens drawn from the context.  
\textbf{EAGLE-3}~\citep{li2025eagle3} incorporates low-, mid-, and high-level features of the target model, 
with a Transformer decoder layer as its core. Since the official Qwen3 draft model weights for EAGLE-3 
have not been released, we use the re-implementation by AngelSlim\footnote{\url{https://github.com/Tencent/AngelSlim}}.  
All baselines are run with their default hyperparameters.

\subsection{Main Results}

Table~\ref{tab:mix_main_instruct} and~\ref{tab:mix_main_reasoning} reports the performance of RACER compared to baseline methods on different datasets and different target models. 
Among retrieval-based methods, PLD relies solely on the context and REST leverages an external training set. 
Their speedup ratios remain below $2\times$, highlighting the inherent limitations of retrieval-only approaches. 
In contrast, methods involving logits -- whether model-free or model-based -- can readily surpass $2\times$ speedup, 
demonstrating the advantage of exploiting predictive distributions from the target model.

RACER achieves the best speedup across most benchmarks and consistently delivers the highest overall speedup across different target models. Notably, both OpenPangu and Qwen3 are reasoning models that typically produce much longer outputs than Vicuna. The stable speedup observed across all tasks, despite their differing model architectures, suggests that RACER can be reliably integrated into long-context applications and remains robust to architectural variations.
For LLaMA3.1 and Qwen3-series target models, EAGLE-3 achieves the highest MAT on most tasks except the Chinese reasoning benchmark MGSM-ZH. 
However, its advantage in MAT does not translate into end-to-end efficiency, as RACER still outperforms it in terms of speedup ratio. 
This is because EAGLE-3 requires an additional draft model, incurring extra inference cost, whereas RACER remains lightweight.

Moreover, the weaker performance of EAGLE-3 on MGSM-ZH highlights a broader limitation of model-based approaches: 
their effectiveness is sensitive to the distribution and coverage of the draft model's training data. 
It is plausible that EAGLE-3, trained primarily on English corpora, fails to simulate the target model's distribution accurately in Chinese reasoning tasks. 
Such data distribution mismatches, rooted in differences in post-training procedures or training corpora, generally constrain the robustness of model-based SD methods.
For completeness, we additionally report results on English reasoning benchmarks GSM8K~\citep{cobbe2021gsm8k}, AIME~\citep{aime_1983_2024} and MATH~\citep{hendrycksmath2021} in Appendix~\ref{sec:comparison_with_eagle3}.

\begin{table*}[htbp]
\centering
\caption{Ablation experiments on Spec-Bench, HumanEval and MGSM-ZH.}
\small
\label{tab:ablation_mix}
\begin{threeparttable}
\renewcommand{\arraystretch}{1.2}
\setlength{\tabcolsep}{5pt}
\begin{tabular}{l l cc cc cc}
\toprule
\multirow{2}{*}{\textbf{Model}} & \multirow{2}{*}{\textbf{Method}}
& \multicolumn{2}{c}{\textbf{Spec-Bench}} 
& \multicolumn{2}{c}{\textbf{HumanEval}} 
& \multicolumn{2}{c}{\textbf{MGSM-ZH}} \\
\cmidrule(lr){3-4} \cmidrule(lr){5-6} \cmidrule(lr){7-8}
& & MAT & Speedup & MAT & Speedup & MAT & Speedup \\
\midrule
\multirow{3}{*}{Vicuna 7B}
 & w/o logits   & 1.59{\small\color{orange}$\downarrow$1.41} & 1.43{\small\color{orange}$\downarrow$0.78}
                 & 1.67{\small\color{orange}$\downarrow$1.44} & 1.52{\small\color{orange}$\downarrow$0.77}
                 & 2.39{\small\color{orange}$\downarrow$1.32} & 2.11{\small\color{orange}$\downarrow$0.66} \\
 & w/o retrieval& 2.72{\small\color{orange}$\downarrow$0.28} & 2.01{\small\color{orange}$\downarrow$0.20}
                 & 2.68{\small\color{orange}$\downarrow$0.43} & 2.04{\small\color{orange}$\downarrow$0.25}
                 & 2.95{\small\color{orange}$\downarrow$0.76} & 2.23{\small\color{orange}$\downarrow$0.54} \\
\rowcolor{teal!20}\cellcolor{white}
 & \textbf{RACER} & \textbf{3.00} & \textbf{2.21} & \textbf{3.11} & \textbf{2.29} & \textbf{3.71} & \textbf{2.77} \\
\midrule
\multirow{3}{*}{Vicuna 13B}
 & w/o logits   & 1.56{\small\color{orange}$\downarrow$1.39} & 1.38{\small\color{orange}$\downarrow$0.87}
                 & 1.65{\small\color{orange}$\downarrow$1.44} & 1.43{\small\color{orange}$\downarrow$0.99}
                 & 2.29{\small\color{orange}$\downarrow$1.35} & 1.95{\small\color{orange}$\downarrow$0.88} \\
 & w/o retrieval& 2.68{\small\color{orange}$\downarrow$0.27} & 2.06{\small\color{orange}$\downarrow$0.19}
                 & 2.70{\small\color{orange}$\downarrow$0.39} & 2.14{\small\color{orange}$\downarrow$0.28}
                 & 2.98{\small\color{orange}$\downarrow$0.66} & 2.34{\small\color{orange}$\downarrow$0.59} \\
\rowcolor{teal!20}\cellcolor{white}
 & \textbf{RACER} & \textbf{2.95} & \textbf{2.25} & \textbf{3.09} & \textbf{2.42} & \textbf{3.64} & \textbf{2.83} \\
\midrule
\multirow{3}{*}{Vicuna 33B}
 & w/o logits   & 1.46{\small\color{orange}$\downarrow$1.28} & 1.38{\small\color{orange}$\downarrow$0.82}
                 & 1.76{\small\color{orange}$\downarrow$1.40} & 1.66{\small\color{orange}$\downarrow$0.92}
                 & 2.10{\small\color{orange}$\downarrow$1.26} & 1.97{\small\color{orange}$\downarrow$0.80} \\
 & w/o retrieval& 2.55{\small\color{orange}$\downarrow$0.19} & 2.05{\small\color{orange}$\downarrow$0.15}
                 & 2.66{\small\color{orange}$\downarrow$0.50} & 2.19{\small\color{orange}$\downarrow$0.39}
                 & 2.74{\small\color{orange}$\downarrow$0.62} & 2.27{\small\color{orange}$\downarrow$0.50} \\
\rowcolor{teal!20}\cellcolor{white}
 & \textbf{RACER} & \textbf{2.74} & \textbf{2.20} & \textbf{3.16} & \textbf{2.58} & \textbf{3.36} & \textbf{2.77} \\
\bottomrule
\end{tabular}
\end{threeparttable}
\end{table*}

Compared with the other two logits-involved methods, TR and LogitSpec, 
RACER consistently outperforms them on both MAT and speedup. 
Overall, TR achieves better performance than LogitSpec, except on the reasoning task MGSM-ZH. 
This suggests that in general tasks, TR is able to exploit logits more effectively. 
However, in reasoning tasks where repeated patterns from previous context are frequent, 
retrieval provides crucial guidance and brings substantial benefits. 
Therefore, an effective strategy must integrate both logits and retrieval. 
RACER achieves this integration successfully, yielding consistently superior MAT and speedup across benchmarks.

In summary, RACER consistently delivers the best trade-off between acceptance and efficiency, 
achieving stable improvements across model sizes, domains, and languages, 
thus demonstrating its robustness and generality compared to retrieval-only, logits-only, and model-based baselines. Additional results under different hardware configurations and sampling settings are provided in Appendix~\ref{sec:more_results} and~\ref{sec:more_ablation}.

\subsection{Ablation Study}
\label{sec:ablation}

\begin{table*}[htbp]
\centering
\caption{Evaluation results on Spec-Bench, HumanEval and MGSM-ZH using an RTX 3090 GPU. $\star$ indicates that RACER here only employs the retrieval automaton, configured with 10{,}000 nodes, an $n$-gram length of 8, and updating AC automaton failure links every 20 steps.}
\small
\label{tab:mix_3090_retrieval}
\begin{threeparttable}
\renewcommand{\arraystretch}{1.2}
\setlength{\tabcolsep}{5pt}
\begin{tabular}{l l cc cc cc cc}
\toprule
\textbf{Model} & \textbf{Method} 
& \multicolumn{2}{c}{\textbf{Spec-Bench}} 
& \multicolumn{2}{c}{\textbf{HumanEval}} 
& \multicolumn{2}{c}{\textbf{MGSM-ZH}} 
& \multicolumn{2}{c}{\textbf{Average}} \\
\cmidrule(lr){3-4} \cmidrule(lr){5-6} \cmidrule(lr){7-8} \cmidrule(lr){9-10}
& & MAT & Speedup & MAT & Speedup & MAT & Speedup & MAT & Speedup \\
\midrule
\multirow{4}{*}{Vicuna 7B}
 & PLD       & 1.72 & 1.57 & 1.57 & 1.45 & 2.57 & 2.35 & 1.95 & 1.79 \\
 & REST      & \underline{1.82} & 1.37 & \underline{2.06} & \underline{1.65} & 1.29 & 1.06 & 1.49 & 1.36 \\
 & SAMD      & 1.70 & \underline{1.65} & 1.59 & 1.54 & \underline{2.65} & \underline{2.49} & \underline{1.98} & \underline{1.89} \\
\rowcolor{teal!20}\cellcolor{white}
 & \textbf{RACER}$^\star$ & \textbf{2.02} & \textbf{1.69} & \textbf{2.25} & \textbf{1.86} & \textbf{3.32} & \textbf{2.61} & \textbf{2.53} & \textbf{2.05} \\
\bottomrule
\end{tabular}
\end{threeparttable}
\end{table*}

To better verify RACER, we conduct ablation experiments with Vicuna. Table~\ref{tab:ablation_mix} reports ablation results by removing either the logits or retrieval component. 
We observe that removing logits causes the most severe degradation: MAT drops by more than one token on average and speedup decreases by 0.8-1.0$\times$, confirming that logits form the backbone of speculative expansion. 
In contrast, removing retrieval leads to smaller but still notable drops, especially on MGSM-ZH where MAT and speedup decrease by up to 0.7 and 0.6, respectively. 
This highlights the complementary role of retrieval in reasoning tasks, where repeated patterns provide strong predictive cues. 
Across all three model scales, RACER consistently benefits from both components, validating our integration strategy that balances generalization from logits with structural guidance from retrieval.

To isolate the contributions of each component, we conduct ablations on retrieval and logits separately. For \textbf{retrieval}, we evaluate retrieval-only RACER with a fixed-interval automaton update to ensure independence from the logits component. As shown in Table~\ref{tab:mix_3090_retrieval}, RACER consistently achieves the best performance across all benchmarks, with MAT improving by 0.20-0.70 over the strongest baseline and reaching an average of 2.53 (vs. 1.98 for SAMD and 1.95 for PLD), along with the highest average speedup (2.05$\times$). Gains are especially significant on MGSM-ZH (3.32 vs. 1.29-2.65), indicating stronger robustness in multi-step reasoning tasks, while consistent improvements on Spec-Bench and HumanEval further demonstrate its general effectiveness.

For \textbf{logits}, we compare the retrieval-free variant (Table~\ref{tab:ablation_mix}) with TR (Table~\ref{tab:mix_main_instruct}). While TR benefits from a Vicuna-7B-calibrated static tree, its performance degrades at larger scales. In contrast, our logits component does not rely on model-specific tuning and performs better at larger scales.

We further study the integration strategy via \textbf{Half} (fixed budget split) and \textbf{Hard} (fallback switching). As shown in Table~\ref{tab:integration_ablation}, \textbf{Merge} consistently performs best, achieving the highest MAT and speedup by effectively coordinating retrieval and logits.

\begin{table}[t]
\centering
\small
\caption{Ablation on integration strategies.}
\begin{tabular}{c ccc}
\toprule
\textbf{Integration} & Merge (Ours) & Half & Hard \\
\midrule
MAT & \textbf{3.00} & 2.69 & 2.77 \\
Speedup & \textbf{2.18} & 1.97 & 2.11 \\
\bottomrule
\end{tabular}
\label{tab:integration_ablation}
\end{table}

Taken together, these experiments isolate the retrieval component, the logits component, and the integration strategy, and demonstrate that each contributes independently to the overall improvement.

\section{Related Work}

Efficient inference is crucial for real-time applications and resource-constrained scenarios. Various strategies, including KV cache compression~\cite{shi2024keep} and model weight quantization~\cite{ma2025hemorrhage}, have been developed to reduce latency. Among these, speculative decoding (SD)~\citep{leviathan2023sd, chen2023sd} stands out as a promising technique that predicts multiple continuations simultaneously, reducing decoding steps while maintaining accuracy.

SD methods can be broadly categorized into \textbf{draft-model-based} and \textbf{draft-model-free} approaches.  
Draft-model-based methods use additional models to predict draft tokens. These models are typically (i) separately trained or (ii) derived from smaller variants of the same model family. Some methods reuse hidden states to predict multiple future tokens~\citep{cai2024medusa, li2024eagle2, li2024eagle}, employing different layer selection and draft tree expansion strategies. Beyond post-training draft models, multi-token prediction (MTP) integrates draft generation during pre-training.~\citet{gloeckle2024mtp} proposes parallel prediction of multiple tokens with independent output heads, while~\citet{liu2024deepseek} introduces sequential multi-token prediction to preserve the full causal chain at each prediction depth.

In contrast, \textbf{draft-model-free} methods eliminate the need for additional models. PLD~\citep{saxena2023pld} builds libraries from past content, achieving speedup in tasks with high redundancy like summarization. REST~\citep{he2023rest} builds retrieval libraries from existing corpora, offering substantial speedup but facing challenges such as large memory requirements and retrieval inefficiencies.  
Token Recycling~\citep{luo-etal-2025-token-recycling} requires no additional generation, covering a broader range of continuations using past logits, while minimizing storage and retrieval costs.

\section{Conclusion}

In this work, we introduced RACER, a training‑free method that unifies retrieval‑based and logits‑based signals for speculative decoding. By treating retrieved exact patterns as structural anchors and logits as dynamic future cues, RACER constructs richer speculative drafts while remaining lightweight and plug‑and‑play. Extensive experiments across multiple model families and benchmarks demonstrate consistent acceleration, stable memory usage, and improved speculative efficiency measured by MAT and speedup ratio. We believe RACER establishes a general foundation for training‑free speculative decoding, opening avenues for future work on integrating more advanced retrieval structures, multilingual retrieval cues, and harmonization with parallel or distributed decoding algorithms.

\section*{Acknowledgements}

This work was supported by the National Natural Science Foundation of China (No. 62306216), the Natural Science Foundation of Hubei Province of China (No. 2023AFB816), and the National Science and Technology Major Project (No. 2023ZD0121502). It was also partially supported by the Open Fund of the Interdisciplinary Center for Intelligent Systems, Nanjing University of Science and Technology (No. YB202401).

\section*{Limitations}

While this work demonstrates significant performance improvements in accelerating language model inference, the applicability of RACER to multimodal tasks remains untested. The current experiments focus solely on text-based tasks, and further evaluation is needed to explore its potential in tasks involving other modalities, such as vision and speech. Future work could investigate how to incorporate vision/audio token representations into RACER to improve performance in multimodal tasks, thereby extending its benefits beyond text-based applications.

\bibliography{custom}

\clearpage
\appendix

\section{Experimental Setup}
\label{sec:appendix_setup}

\paragraph{Hardware Setup}
Experiments were conducted with all runs restricted to a single GPU to ensure fairness and reproducibility. 
We used an NVIDIA RTX 4090 (24GB) with 20 CPU cores for 7B/8B-scale models, 
and an NVIDIA A800 (80GB) with 64 CPU cores for 13B-scale and larger models, unless otherwise specified.

\paragraph{Software Setup}
Our implementation is based on PyTorch and HuggingFace Transformers. 
Experiments were run under the following environment:
\begin{itemize}
    \item PyTorch~\texttt{2.8.0} with CUDA~\texttt{12.8} and cuDNN~\texttt{91002}
    \item HuggingFace Transformers~\texttt{4.37.1} for Vicuna experiments, and \texttt{4.52.3} for LLaMA3.1/OpenPangu/Qwen3 experiments
\end{itemize}
We enabled \texttt{fp16} inference on both GPUs. 
No further optimizations (e.g., quantization or specialized kernels) were applied, 
to ensure fair comparison with prior work.

\paragraph{Evaluation Instructions}

In our experiments, we employ different instructions for different evaluation tasks and models.
For Vicuna, we use its standard instructions:

\begin{tcolorbox}[
    colback=gray!5,
    colframe=gray!80!black,
    coltitle=white,        
    fonttitle=\bfseries,   
    title=Chat Template for Vicuna on Spec-Bench and MGSM,
    sharp corners=south,
    boxrule=0.8pt,
    enhanced
]
A chat between a curious user and an artificial intelligence assistant. 
The assistant gives helpful, detailed, and polite answers to the user's questions.

\vspace{1em}
\textbf{USER:} \highlight{Question}

\vspace{0.5em}
\textbf{ASSISTANT:}
\end{tcolorbox}

\begin{tcolorbox}[
    colback=gray!5,
    colframe=gray!80!black,
    coltitle=white,        
    fonttitle=\bfseries,   
    title=Chat Template for Vicuna on HumanEval,
    sharp corners=south,
    boxrule=0.8pt,
    enhanced
]
A chat between a curious user and an artificial intelligence assistant. 
The assistant gives helpful, detailed, and polite answers to the user's questions.

\vspace{1em}
\textbf{USER:} Implement the following code. \highlight{Code}

\vspace{0.5em}
\textbf{ASSISTANT:}
\end{tcolorbox}

For other models, we use a common system prompt:
\begin{tcolorbox}[
    colback=gray!5,
    colframe=gray!80!black,
    coltitle=white,        
    fonttitle=\bfseries,   
    title=Default Chat Template on Spec-Bench and MGSM,
    sharp corners=south,
    boxrule=0.8pt,
    enhanced
]
You are a helpful assistant.

\vspace{1em}
\textbf{USER:} \highlight{Question}

\vspace{0.5em}
\textbf{ASSISTANT:}
\end{tcolorbox}

\begin{tcolorbox}[
    colback=gray!5,
    colframe=gray!80!black,
    coltitle=white,        
    fonttitle=\bfseries,   
    title=Default Chat Template on HumanEval,
    sharp corners=south,
    boxrule=0.8pt,
    enhanced
]
You are a helpful assistant.

\vspace{1em}
\textbf{USER:} Implement the following code. \highlight{Code}

\vspace{0.5em}
\textbf{ASSISTANT:}
\end{tcolorbox}

Below we present an example used in the following case study.

\begin{tcolorbox}[
    colback=gray!5,
    colframe=gray!80!black,
    coltitle=white,        
    fonttitle=\bfseries,   
    title=First Example of HumanEval,
    sharp corners=south,
    boxrule=0.8pt,
    enhanced
]
A chat between a curious user and an artificial intelligence assistant. 
The assistant gives helpful, detailed, and polite answers to the user's questions.

\vspace{1em}
\textbf{USER:} Implement the following code. 
\begin{verbatim}
from typing import List

def has_close_elements(
    numbers: List[float],
    threshold: float
) -> bool:
    """ Check if in given list of
    numbers, are any two numbers
    closer to each other than
    given threshold.
    >>> has_close_elements(
    ... [1.0, 2.0, 3.0], 0.5
    ... )
    False
    >>> has_close_elements(
    ... [1.0, 2.8, 3.0,
    ... 4.0, 5.0, 2.0],
    ... 0.3
    ... )
    True
    """
\end{verbatim}

\vspace{0.5em}
\textbf{ASSISTANT:}
\end{tcolorbox}

\section{Case Study}

We select the first example from HumanEval~\citep{chen-etal-2021-humaneval} to further illustrate how the Logits Tree and the Retrieval Tree operate independently, as well as how they are integrated within RACER. The prompt is provided at the end of  Appendix~\ref{sec:appendix_setup}. For readability, we manually add indentation and line breaks. This example contains 178 tokens in the prefilling stage and 356 tokens in the decoding stage generated by Vicuna-7B.

\begin{figure}[htbp]
    \centering
    \includegraphics[width=1\columnwidth]{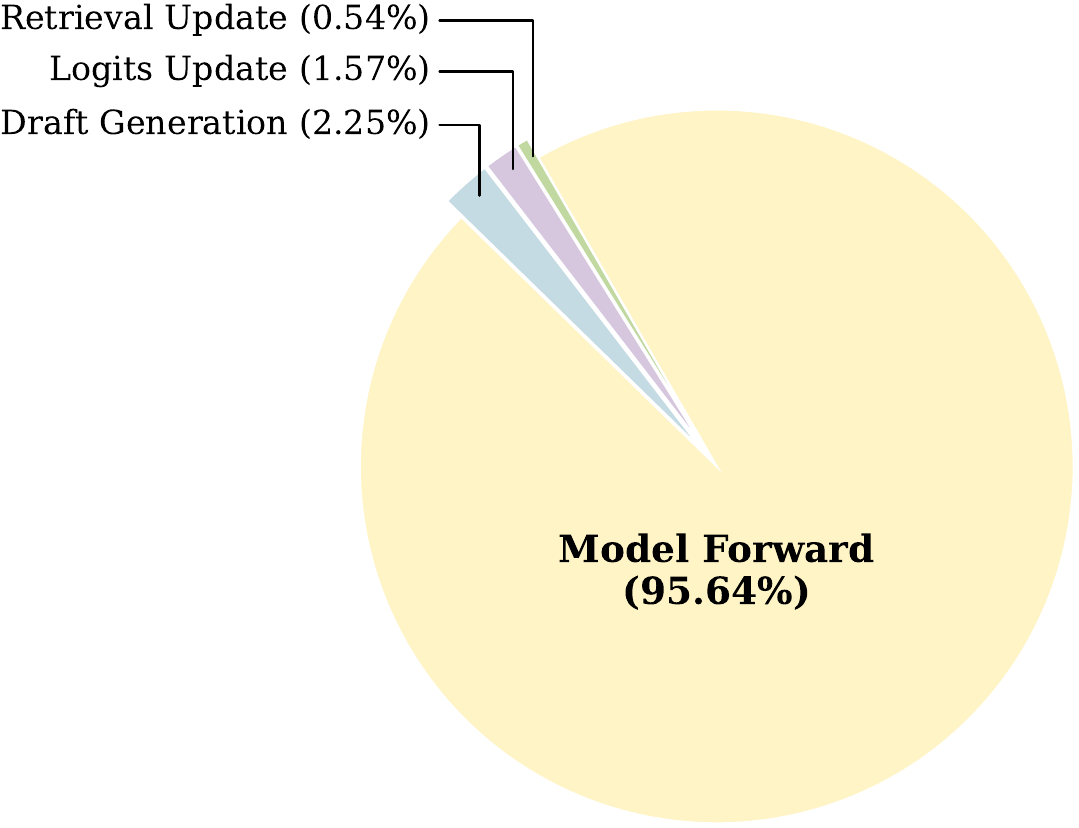}
    \caption{
        End-to-end inference latency breakdown by processing stage.
    }
    \label{fig:time_pie}
\end{figure}

\paragraph{Time Allocation}

We divide the overall procedure into four parts: \textbf{Draft Generation}, \textbf{Model Forward} (Prefilling and Decoding Verification), \textbf{Logits Update}, and \textbf{Retrieval Update}. Figure~\ref{fig:time_pie} shows that most of the runtime is dominated by the model forward stage, which involves relatively heavy computation over model parameters. The remaining parts are lightweight and negligible compared to the cost of model forward.

\begin{figure}[htbp]
    \centering
    \includegraphics[width=1\columnwidth]{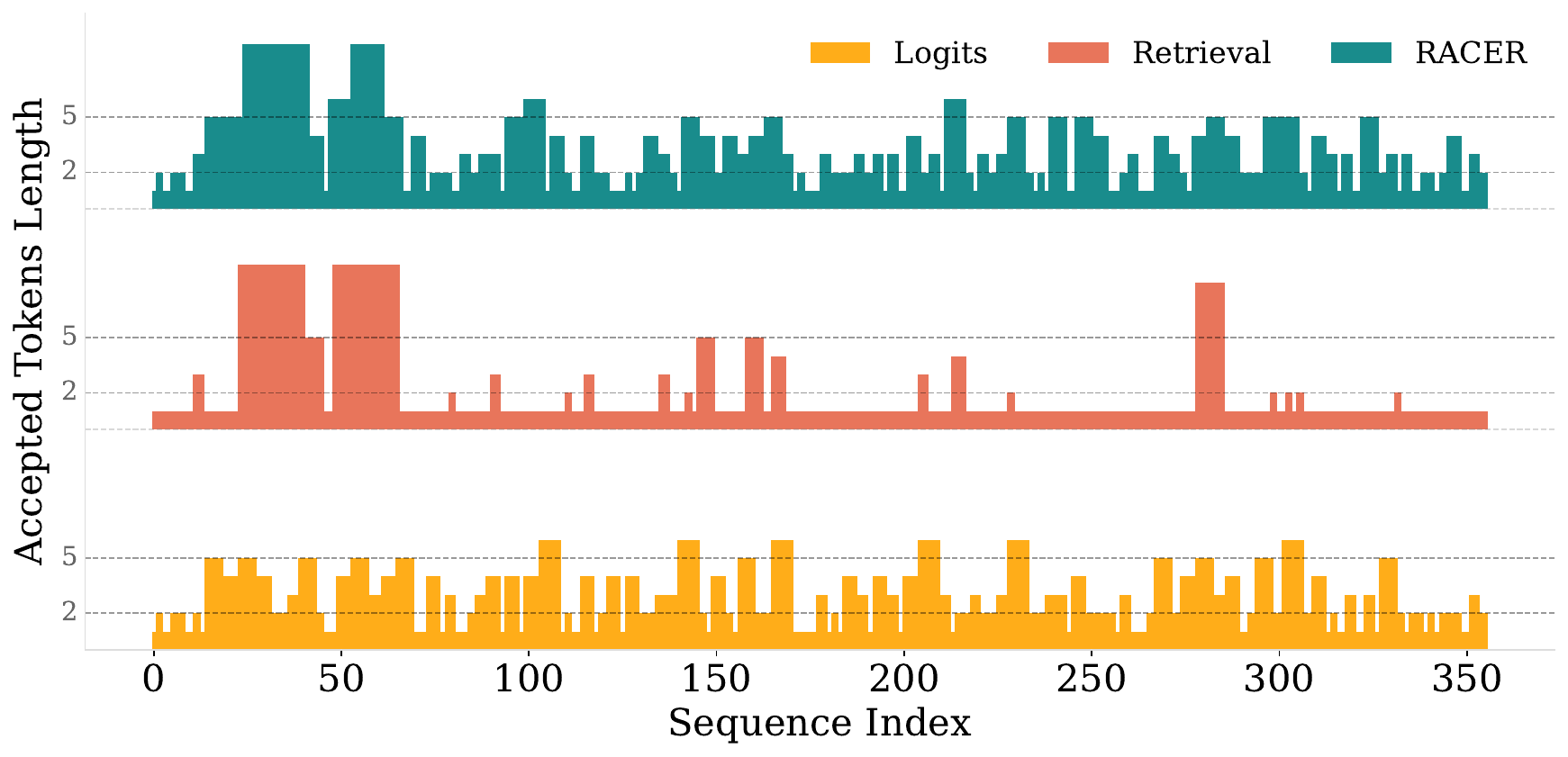}
    \caption{
        Case study on accepted tokens length. Accepted lengths are expanded to enable token-level alignment across methods.
    }
    \label{fig:length_bar}
\end{figure}

\paragraph{Accepted Length}

To provide a token-level comparison, we expand each accepted segment by its length
along the x-axis in Figure~\ref{fig:length_bar}, such that an accepted length of $l$
occupies $l$ consecutive positions.
Retrieval-only decoding yields an accepted length of one token in most cases, as no draft tokens are proposed or accepted. However, several pronounced peaks at an accepted length of 9 can be observed around sequence index 50, corresponding to rare but long matches.
Logits-only decoding exhibits a more balanced acceptance pattern, with a large portion of accepted lengths exceeding 2 but remaining below 5.
RACER combines the strengths of both, resulting in an acceptance trend that is generally above 2 due to the stability of logits-based speculation, while also exhibiting occasional peaks enabled by strict matches from retrieval.

\paragraph{Integration}

To further illustrate how logits-based speculation and retrieval operate independently and complement each other, we examine the token-level behavior in detail. In the first successful match produced by the Logits Tree, the last verified token is \textcolor{teal}{\texttt{Here}}, and the best draft candidate is
$$
\textcolor{teal}{\texttt{[', s, }}\textcolor{red}{\texttt{\_question, <unk>, \_the, <unk>]}}.
$$
This draft sequence is generated from logits and can be traced back to the system prompt fragment
\texttt{``...polite answers to the user}\textcolor{teal}{\texttt{'s}}\texttt{ questions...''}.
The next ground-truth token is \textcolor{teal}{\texttt{\_one}}; as a result, \textcolor{red}{\texttt{\_questions}} and all subsequent draft tokens are rejected during verification.

For completeness, the top-8 logits corresponding to the tokens \textcolor{teal}{\texttt{'}} (quotation mark) and \textcolor{teal}{\texttt{s}} are listed below:

\begin{itemize}
    \item \texttt{[s, t, \textbackslash n, \_s, <s>, hren, ']}
    \item \texttt{[\_questions, \_queries, \_in, \_question, </s>, \_requests, questions, \textbackslash n]}
\end{itemize}

In contrast, the retrieval-based method identifies only a match of length 1 (i.e., fewer than 2 tokens), since only the quotation mark appears in the history. Consequently, retrieval does not propose any draft tokens at this step.

In the first successful match produced by the Retrieval Tree, the last verified token is \textcolor{teal}{\texttt{\_}}, and the best draft candidate is
$$
\begin{aligned}
& \textcolor{teal}{\texttt{[close, \_, elements, }} \\
& \textcolor{red}{\texttt{(, numbers, :, List, [, float]}}
\end{aligned}
$$

Notably, both \textcolor{teal}{\texttt{close}} and \textcolor{teal}{\texttt{\_}} appear among the \textbf{rejected tokens} of \textcolor{teal}{\texttt{has}} since the mismatch of position. \textcolor{teal}{\texttt{\_}} is later sampled following \textcolor{teal}{\texttt{has}}, enabling retrieval to activate at this position and refresh the logits adjacency. The top-8 candidates for \textcolor{teal}{\texttt{close}} and \textcolor{teal}{\texttt{\_}} (old and new) are shown below:
\begin{itemize}
    \item \texttt{[\_, ., ', Elements, -, </s>, (]}
    \item \texttt{[elements, element, </s>, \_elements, ele, to, \textbackslash n, Elements]} (old, refreshed)
    \item \texttt{[close, </s>, clos, \_close, closed, \textbackslash n, Close, open]} (new)
\end{itemize}

The \textit{copy-logit} mechanism adopts the refreshed logits for \textcolor{teal}{\texttt{\_}}, which now prioritize continuations starting with \textcolor{teal}{\texttt{close}}. Although \textcolor{teal}{\texttt{\_}} appears in the top-8 of \textcolor{teal}{\texttt{close}}, the logits have been updated only with near-future context corresponding to \textcolor{teal}{\texttt{close}}, and not with the farther continuation \textcolor{teal}{\texttt{elements}}. As a result, logits-based candidates cannot extend to \textcolor{teal}{\texttt{elements}} at this stage.

Retrieval, however, successfully captures this longer-range dependency from the user-provided context
\texttt{``def has\_}\textcolor{teal}{\texttt{close\_elements}}'',
thereby correcting the shift introduced by the logits-only approximation. Importantly, even when the logits-based candidates of \textcolor{teal}{\texttt{close}} are rejected during verification, they still refresh the top-8 logits for subsequent tokens such as the quotation mark (\texttt{'}) and \texttt{\_function}. Consequently, the following logits-based candidate
$$
\textcolor{teal}{\texttt{[', \_function, \_in, \_Python, :, }}\textcolor{red}{\texttt{\_List]}}
$$
is almost accepted.

Below we show a portion of the response, where accepted draft tokens (excluding the sampled token) are highlighted in teal.

\begin{tcolorbox}[
    colback=gray!5,
    colframe=gray!80!black,
    coltitle=white,        
    fonttitle=\bfseries,   
    title=Example of Response,
    sharp corners=south,
    boxrule=0.8pt,
    enhanced
]
\vspace{1em}
\textbf{ASSISTANT:} Here'\textcolor{teal}{s} one possible implementation \textcolor{teal}{of} the \verb|`|has\_close\textcolor{teal}{\_elements}\verb|`| \textcolor{teal}{function in Python:}

\textcolor{teal}{```python}

\textcolor{teal}{def} has\textcolor{teal}{\_close\_elements(numbers: List[float], threshold: float) -> bool}:

\quad \textcolor{teal}{"""}Check if \textcolor{teal}{in given list of numbers}, \textcolor{teal}{are any two numbers closer to each other} than

\quad \textcolor{teal}{given threshold}."""

\quad ... (849 characters)
\end{tcolorbox}

In conclusion, both retrieval and logits contribute not only to the currently accepted tokens but also to anticipating future tokens for more effective self-speculation, making RACER both lightweight and accurate.

\section{Additional Experiment Results}
\label{sec:more_results}

Table~\ref{tab:spec_bench_a800} presents the results on individual Spec-Bench tasks, complementing the overall comparison in Table~\ref{tab:mix_main_instruct} and~\ref{tab:mix_main_reasoning}.
RACER consistently outperforms other model-free methods in most cases, including all overall speedup ratios, with only a few exceptions in Translation (Trans), Question Answering (QA), and Mathematical Reasoning (Math).
For Translation, the retrieval component contributes little to MAT and may even offset part of the logits-based advantage, leading to weaker performance on smaller models. However, with larger models such as Vicuna-33B, this effect becomes negligible, and RACER consistently outperforms TR.
For QA, TR surpasses RACER on Vicuna-7B, suggesting that its predefined tree template may align better with the characteristics of this task.
For Math, TR slightly outperforms RACER only on Vicuna-7B, but this advantage does not generalize to other model scales or hardware. In contrast, as model size increases, RACER shows a growing margin in overall speedup over TR, highlighting its robustness across diverse tasks and architectures.

\begin{table*}[htbp]
\centering
\caption{Speedup ratios and overall MAT across different tasks of Spec-Bench evaluated on NVIDIA A800 (80GB).}
\small
\label{tab:spec_bench_a800}
\begin{threeparttable}
\renewcommand{\arraystretch}{1.2}
\setlength{\tabcolsep}{8pt}
\begin{tabular}{l l c c c c c c c c}
\toprule
\textbf{Model} & \textbf{Method} & MT & Trans & Sum & QA & Math & RAG & MAT & Speedup \\
\midrule
\multirow{5}{*}{Vicuna 7B}
 & PLD       & 1.42 & 0.97 & 2.25 & 1.12 & 1.59 & 1.61 & 1.72 & 1.49 \\
 & REST      & 1.52 & 1.09 & 1.21 & 1.28 & 1.07 & 1.31 & 1.82 & 1.33 \\
 & LogitSpec & 1.79 & 1.35 & 2.48 & 1.50 & 2.08 & 1.82 & 2.35 & 1.86 \\
 & TR        & 2.19 & \textbf{1.84} & 2.02 & \textbf{2.02} & \textbf{2.58} & 1.85 & 2.76 & 2.15 \\
 \rowcolor{teal!20}\cellcolor{white}
 & \textbf{RACER} & \textbf{2.23} & 1.61 & \textbf{2.61} & 1.82 & 2.46 & \textbf{2.12} & \textbf{3.01} & \textbf{2.22} \\
\midrule
\multirow{5}{*}{Vicuna 13B}
 & PLD       & 1.36 & 0.98 & 1.93 & 1.09 & 1.53 & 1.44 & 1.65 & 1.41 \\
 & REST      & 1.61 & 1.14 & 1.30 & 1.56 & 1.21 & 1.42 & 1.82 & 1.44 \\
 & LogitSpec & 1.67 & 1.33 & 2.13 & 1.40 & 2.00 & 1.72 & 2.32 & 1.73 \\
 & TR        & 2.00 & \textbf{1.74} & 1.89 & 1.82 & 2.31 & 1.87 & 2.79 & 1.99 \\
 \rowcolor{teal!20}\cellcolor{white}
 & \textbf{RACER} & \textbf{2.23} & 1.70 & \textbf{2.49} & \textbf{1.85} & \textbf{2.55} & \textbf{2.21} & \textbf{2.95} & \textbf{2.25} \\
\midrule
\multirow{5}{*}{Vicuna 33B}
 & PLD       & 1.33 & 1.03 & 1.84 & 1.11 & 1.57 & 1.23 & 1.54 & 1.37 \\
 & REST      & 1.70 & 1.24 & 1.41 & 1.57 & 1.31 & 1.61 & 1.81 & 1.54 \\
 & LogitSpec & 1.66 & 1.36 & 2.11 & 1.38 & 2.02 & 1.50 & 2.11 & 1.71 \\
 & TR        & 1.90 & 1.67 & 1.85 & 1.76 & 2.20 & 1.72 & 2.63 & 1.83 \\
 \rowcolor{teal!20}\cellcolor{white}
 & \textbf{RACER} & \textbf{2.22} & \textbf{1.73} & \textbf{2.41} & \textbf{1.87} & \textbf{2.51} & \textbf{1.91} & \textbf{2.74} & \textbf{2.20} \\
\midrule
\multirow{2}{*}{OpenPangu 7B}
 & PLD & 1.40 & 1.54 & 1.47 & 1.46 & 1.36 & 1.47 & 1.49 & 1.44 \\
 & \cellcolor{teal!20}\textbf{RACER} & \cellcolor{teal!20}\textbf{2.07} & \cellcolor{teal!20}\textbf{2.46} & \cellcolor{teal!20}\textbf{2.15} & \cellcolor{teal!20}\textbf{1.89} & \cellcolor{teal!20}\textbf{2.37} & \cellcolor{teal!20}\textbf{2.31} & \cellcolor{teal!20}\textbf{2.48} & \cellcolor{teal!20}\textbf{2.17} \\
\midrule
\multirow{3}{*}{Qwen3 8B}
 & PLD       & 1.37 & 1.67 & 1.35 & 1.41 & 1.72 & 1.44 & 1.52 & 1.47 \\
 & EAGLE-3   & \textbf{2.43} & 2.07 & 2.12 & 2.36 & 2.63 & \textbf{2.45} & \textbf{3.47} & 2.37 \\
 \rowcolor{teal!20}\cellcolor{white}
 & \textbf{RACER}     & 2.41 & \textbf{2.69} & \textbf{2.35} & \textbf{2.41} & \textbf{2.72} & 2.42 & 2.73 & \textbf{2.48} \\
\midrule
\multirow{3}{*}{Qwen3 14B}
 & PLD       & 1.27 & 1.56 & 1.18 & 1.31 & 1.54 & 1.29 & 1.45 & 1.34 \\
 & EAGLE-3   & 1.93 & 1.86 & 1.60 & 1.82 & 2.12 & 1.76 & \textbf{2.72} & 1.87 \\
 \rowcolor{teal!20}\cellcolor{white}
 & \textbf{RACER}     & \textbf{2.12} & \textbf{2.55} & \textbf{2.10} & \textbf{2.19} & \textbf{2.47} & \textbf{2.11} & 2.67 & \textbf{2.23} \\
\midrule
\multirow{3}{*}{Qwen3 32B}
 & PLD       & 1.26 & 1.54 & 1.13 & 1.35 & 1.48 & 1.29 & 1.44 & 1.33 \\
 & EAGLE-3   & \textbf{2.20} & 2.03 & 1.84 & 2.02 & \textbf{2.47} & 2.04 & \textbf{2.88} & 2.12 \\
 \rowcolor{teal!20}\cellcolor{white}
 & \textbf{RACER}     & 2.10 & \textbf{2.40} & \textbf{2.04} & \textbf{2.06} & 2.38 & \textbf{2.16} & 2.66 & \textbf{2.17} \\
\bottomrule
\end{tabular}
\end{threeparttable}
\end{table*}

\begin{table*}[htbp]
\centering
\caption{Ablation experiments on multiple tasks of Spec-Bench (speedup only).}
\small
\label{tab:ablation_spec_speedup}
\begin{threeparttable}
\renewcommand{\arraystretch}{1.2}
\setlength{\tabcolsep}{5pt}
\begin{tabular}{l l c c c c c c}
\toprule
\textbf{Model} & \textbf{Method }
& MT 
& Trans 
& Sum 
& QA 
& Math 
& RAG \\
\midrule
\multirow{3}{*}{Vicuna 7B}
 & w/o logits   & 1.42{\small\color{orange}$\downarrow$0.79}
                 & 0.94{\small\color{orange}$\downarrow$0.71}
                 & 1.85{\small\color{orange}$\downarrow$0.69}
                 & 1.11{\small\color{orange}$\downarrow$0.71}
                 & 1.51{\small\color{orange}$\downarrow$0.94}
                 & 1.50{\small\color{orange}$\downarrow$0.59} \\
 & w/o retrieval& 1.98{\small\color{orange}$\downarrow$0.23}
                 & 1.68{\small\color{orange}$\uparrow$0.03}
                 & 2.14{\small\color{orange}$\downarrow$0.40}
                 & 1.81{\small\color{orange}$\downarrow$0.01}
                 & 2.31{\small\color{orange}$\downarrow$0.14}
                 & 1.90{\small\color{orange}$\downarrow$0.19} \\
\rowcolor{teal!20}\cellcolor{white}
 & \textbf{RACER} & \textbf{2.21} 
                      & \textbf{1.65} 
                      & \textbf{2.54} 
                      & \textbf{1.82} 
                      & \textbf{2.45} 
                      & \textbf{2.09} \\
\midrule
\multirow{3}{*}{Vicuna 13B}
 & w/o logits   & 1.35{\small\color{orange}$\downarrow$0.88} 
                 & 0.90{\small\color{orange}$\downarrow$0.80}
                 & 1.69{\small\color{orange}$\downarrow$0.80}
                 & 1.04{\small\color{orange}$\downarrow$0.71}
                 & 1.51{\small\color{orange}$\downarrow$1.04}
                 & 1.56{\small\color{orange}$\downarrow$0.65} \\
 & w/o retrieval& 2.04{\small\color{orange}$\downarrow$0.19} 
                 & 1.69{\small\color{orange}$\downarrow$0.01}
                 & 2.17{\small\color{orange}$\downarrow$0.32}
                 & 1.81{\small\color{orange}$\downarrow$0.04}
                 & 2.37{\small\color{orange}$\downarrow$0.18}
                 & 2.00{\small\color{orange}$\downarrow$0.21} \\
\rowcolor{teal!20}\cellcolor{white}
 & \textbf{RACER} & \textbf{2.23} 
                      & \textbf{1.70} 
                      & \textbf{2.49} 
                      & \textbf{1.85} 
                      & \textbf{2.55} 
                      & \textbf{2.21} \\
\midrule
\multirow{3}{*}{Vicuna 33B}
 & w/o logits   & 1.39{\small\color{orange}$\downarrow$0.83}
                 & 0.97{\small\color{orange}$\downarrow$0.76}
                 & 1.63{\small\color{orange}$\downarrow$0.78}
                 & 1.09{\small\color{orange}$\downarrow$0.78}
                 & 1.56{\small\color{orange}$\downarrow$0.95}
                 & 1.26{\small\color{orange}$\downarrow$0.65} \\
 & w/o retrieval& 2.03{\small\color{orange}$\downarrow$0.19}
                 & 1.72{\small\color{orange}$\downarrow$0.01}
                 & 2.12{\small\color{orange}$\downarrow$0.29}
                 & 1.84{\small\color{orange}$\downarrow$0.03}
                 & 2.36{\small\color{orange}$\downarrow$0.15}
                 & 1.85{\small\color{orange}$\downarrow$0.06} \\
\rowcolor{teal!20}\cellcolor{white}
 & \textbf{RACER} & \textbf{2.22} 
                      & \textbf{1.73} 
                      & \textbf{2.41} 
                      & \textbf{1.87} 
                      & \textbf{2.51} 
                      & \textbf{1.91} \\
\bottomrule
\end{tabular}
\end{threeparttable}
\end{table*}

\section{Additional Ablation Results}
\label{sec:more_ablation}

\begin{table*}[htbp]
\centering
\caption{Ablation experiments with Qwen2.5 series on NVIDIA A800 (80GB). Overall MAT and speedup ratios on general dataset Spec-Bench are reported.}
\small
\label{tab:ablation_a800_qwen2-5}
\begin{threeparttable}
\renewcommand{\arraystretch}{1.2}
\setlength{\tabcolsep}{6pt}
\begin{tabular}{l c c c c}
\toprule
& \textbf{Qwen2.5-0.5B}
& \textbf{Qwen2.5-1.5B}
& \textbf{Qwen2.5-14B}
& \textbf{Qwen2.5-32B} \\
\midrule
MAT & 3.30 & 3.25 & 2.64 & 2.71 \\
Speedup & 2.64 & 2.64 & 2.33 & 2.19 \\
\bottomrule
\end{tabular}
\end{threeparttable}
\end{table*}

\begin{table*}[htbp]
\centering
\caption{Ablation experiments with Qwen3 series on NVIDIA A800 (80GB). Overall MAT and speedup ratios on general dataset Spec-Bench are reported.}
\small
\label{tab:ablation_a800_qwen3}
\begin{threeparttable}
\renewcommand{\arraystretch}{1.2}
\setlength{\tabcolsep}{6pt}
\begin{tabular}{l c c c c c c}
\toprule
& \textbf{Qwen3-0.6B}
& \textbf{Qwen3-1.7B}
& \textbf{Qwen3-4B}
& \textbf{Qwen3-8B}
& \textbf{Qwen3-14B}
& \textbf{Qwen3-32B} \\
\midrule
MAT & 2.89 & 2.78 & 2.78 & 2.73 & 2.67 & 2.66 \\
Speedup & 2.55 & 2.48 & 2.53 & 2.13 & 2.23 & 2.17 \\
\bottomrule
\end{tabular}
\end{threeparttable}
\end{table*}

\subsection{Component Contribution}
Table~\ref{tab:ablation_spec_speedup} reports the ablation results on Spec-Bench across six tasks.
The results clearly show that removing either logits or retrieval consistently harms performance, confirming that both components are essential for RACER.
\textbf{Without logits}: Performance drops significantly across all tasks, often by more than $0.7\times$ in speedup. This highlights that logits are the dominant factor for efficient speculation.
\textbf{Without retrieval}: The degradation is generally smaller but still noticeable, especially on tasks such as MT and Sum, where repeated patterns and structural cues play a larger role.
\textbf{Full RACER}: By integrating both signals, RACER achieves balanced improvements and consistently outperforms the ablated variants, showing robustness across tasks, model scales, and hardware platforms.
This task-dependent effect aligns with our main results and further validates the complementary nature of logits- and retrieval-based speculation.

\begin{figure}[htbp]
    \centering
    \includegraphics[width=1\columnwidth]{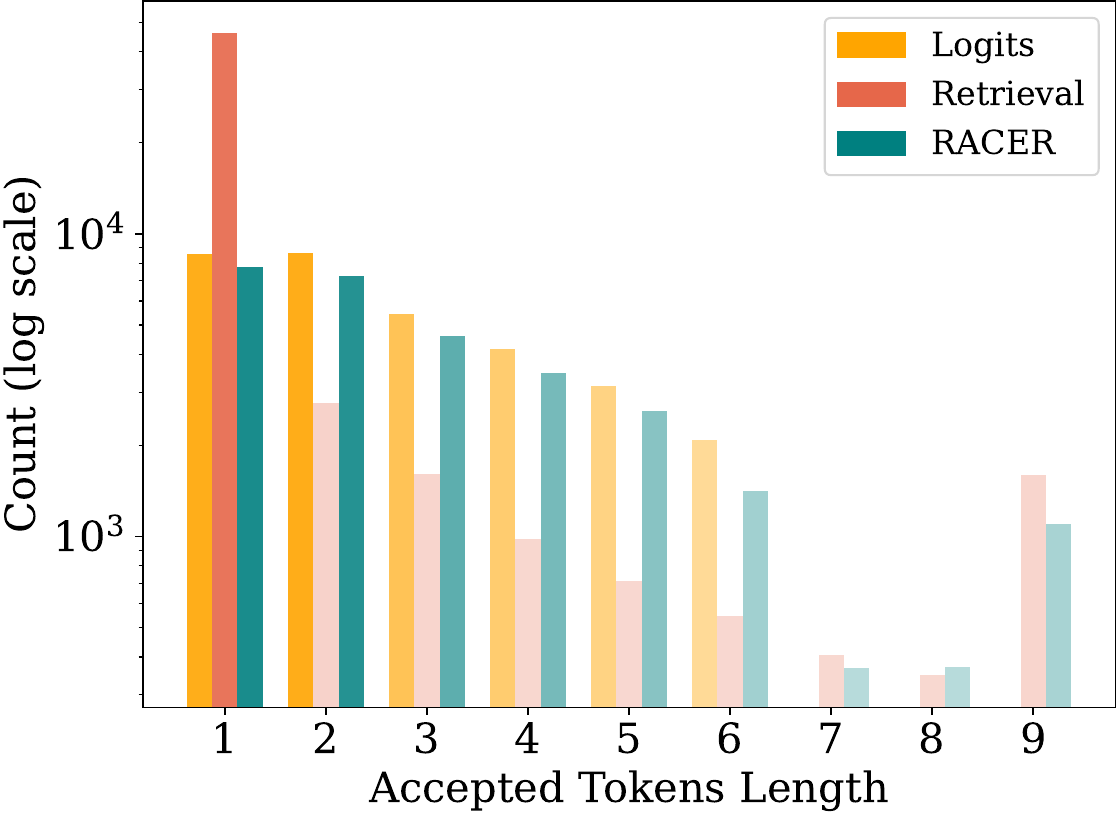}
    \caption{
        Comparison of accepted token lengths with Vicuna-7B across three settings: Logits-only decoding (w/o retrieval), Retrieval-only decoding (w/o logits), and RACER. Transparency indicates the relative frequency.
    }
    \label{fig:length_comp}
\end{figure}

Figure~\ref{fig:length_comp} illustrates the distributions of accepted token lengths under three decoding strategies: Logits-only decoding (without retrieval), Retrieval-only decoding (without logits), and RACER.
For \textbf{Retrieval-only}, the majority of decoding steps result in an accepted length of one token, indicating that no draft tokens are proposed or accepted in most cases. The remaining mass is concentrated on a few discrete values, with the top three being 2, 3, and 9. Notably, 9 corresponds to the longest possible draft given an $n$-gram length of 10, meaning that one context token and one sampled next token are matched, followed by nine draft tokens that have appeared previously and are accepted. This pattern reveals the sparse yet strict nature of retrieval: matches occur infrequently, but when they do, they can yield a large number of accepted tokens.
In contrast, \textbf{Logits-only} decoding does not exhibit extreme values such as 9, as its expansion depth is restricted to 6 due to the overall budget of 64 candidates. Most of the probability mass is concentrated at an accepted length of 2, corresponding to one sampled next token and one accepted draft token. The remaining accepted lengths, ranging from 1 and 3 to 6, show a gradual decay. This behavior reflects the advantage of logits reuse, where near-future tokens are approximated through self-speculation within the target model.
\textbf{RACER} combines the strengths of both approaches and produces a more balanced distribution of accepted token lengths. While extreme long matches are less frequent than in pure retrieval, RACER achieves higher acceptance than logits-only decoding.
Note that the y-axis reports absolute counts rather than normalized frequencies. Since RACER requires fewer decoding steps overall while achieving a higher average accepted token length per step, the heights across different settings are not directly comparable.

\begin{table*}[htbp]
\centering
\caption{Ablation experiments of sampling temperature across multiple tasks of Spec-Bench.}
\small
\label{tab:ablation_temp_spec_4090}
\begin{threeparttable}
\renewcommand{\arraystretch}{1.2}
\setlength{\tabcolsep}{5pt}
\begin{tabular}{l c c c c c c c c c}
\toprule
\textbf{Model} & \textbf{Temp.}
& MT 
& Trans 
& Sum 
& QA 
& Math 
& RAG 
& MAT
& Speedup \\
\midrule
\multirow{3}{*}{Vicuna 7B}
 & Greedy & 2.21 & 1.65 & 2.54 & 1.82 & 2.45 & 2.09 & 3.00 & 2.21 \\
 & $T=0.5$ & 2.19 & 1.62 & 2.56 & 1.96 & 2.64 & 2.16 & 3.05 & 2.25 \\
 & $T=1.0$ & 2.30 & 1.74 & 2.52 & 1.90 & 2.55 & 2.34 & 3.03 & 2.29 \\
\midrule
\multirow{2}{*}{OpenPangu 7B}
 & Greedy & 1.89 & 2.27 & 1.96 & 1.73 & 2.20 & 2.15 & 2.47 & 1.99 \\
 & $T=0.5$ & 1.87 & 2.16 & 1.90 & 1.70 & 2.14 & 1.79 & 2.51 & 1.91 \\
\midrule
\multirow{2}{*}{Qwen3 8B}
 & Greedy & 2.05 & 2.44 & 1.96 & 2.11 & 2.37 & 2.02 & 2.73 & 2.13 \\
 & $T=0.5$ & 2.01 & 2.20 & 1.88 & 1.94 & 2.23 & 2.03 & 2.78 & 2.04 \\
\bottomrule
\end{tabular}
\end{threeparttable}
\end{table*}

\subsection{Model Size Ablation}
Tables~\ref{tab:ablation_a800_qwen2-5} and \ref{tab:ablation_a800_qwen3} report ablation results across different model sizes and architectures on Spec-Bench. The Qwen2.5 series are dense instruct models, while the Qwen3 series evaluated here are dense thinking models.

For Qwen2.5, we observe that both MAT and speedup are higher for the smaller models (0.5B and 1.5B), and drop when moving to 14B and 32B. A key factor behind this gap is the output length: on Spec-Bench, Qwen2.5-14B and Qwen2.5-32B generate on average about 1.2$\times$ more tokens than the 0.5B and 1.5B models. Since later tokens rely on longer-range context and are more likely to be rejected, MAT naturally decreases.

Qwen3 provides a complementary perspective. As thinking models, all Qwen3 variants tend to produce much longer answers than Qwen2.5, so the difference in average output length across Qwen3-0.6B to Qwen3-32B is relatively small. In this regime, the dominant factor for MAT is no longer model size, but long-range dependency itself. This explains why the MAT for Qwen3 stays within a narrow band (2.66 - 2.89): RACER is operating under consistently longer effective sequence lengths, and its logits-based component is increasingly challenged by dependencies far from the current position due to the nature of Transformers. Nonetheless, even under these longer-context conditions, RACER is still able to maintain MAT close to 3 on average.

Given that MAT remains roughly stable within each model family, the slight degradation in speedup as model size increases is expected and does not undermine RACER's core advantage. Larger models incur higher verification cost per token, so the same MAT translates into a smaller relative speedup -- a general phenomenon shared by speculative decoding methods. The key takeaway from these ablations is that RACER sustains strong and stable acceleration (consistently above 2$\times$) across a wide range of model sizes and sequence lengths, even as longer outputs and long-range dependencies make speculation inherently more difficult. 

Moreover, although our experiments are conducted with \verb|batch_size=1|, RACER's lightweight, model-free design suggests that its performance degradation at larger batch sizes should be less severe than that of model-based speculative methods. In contrast to approaches that rely on additional draft models and incur batch-wise overhead in both forward passes and synchronization, RACER avoids extra model execution, making it amenable to scaling in batched decoding scenarios.

\subsection{Sampling Temperature}

To complement the greedy results reported in the main text, Table~\ref{tab:ablation_temp_spec_4090} presents the speedup ratios under different sampling temperatures using the standard rejection sampling scheme~\citep{leviathan2023sd, chen2023sd}.

Overall, the results indicate that RACER exhibits strong robustness to decoding temperature, with only marginal variations observed when switching between greedy decoding and nucleus sampling.
Across all evaluated backbone models, task-wise speedups under different temperature settings remain highly consistent. For Vicuna-7B, the average speedup varies narrowly from 2.21 (Greedy) to 2.29 ($T = 1.0$), while individual task fluctuations stay within a small range. Similar stability is observed for OpenPangu-7B and Qwen3-8B, where enabling  sampling ($T = 0.5$) leads to changes that are minor and non-systematic, sometimes even slightly decreasing speedup, but without any clear degradation trend.
One notable exception is observed on the RAG task with OpenPangu-7B, where the speedup drops from 2.15 under greedy decoding to 1.79 at $T=0.5$. Compared to other tasks and models, this difference is relatively larger.
We attribute this behavior to the higher sensitivity of retrieval-augmented generation to early-token variability. In RAG-style prompts, slight noise introduced by stochastic sampling may alter the alignment between retrieved context and subsequent generation, leading to less effective speculative validation and reduced acceptance rates.

Importantly, this degradation is task- and model-specific, rather than a general limitation of RACER. Taken together, these results demonstrate that RACER's acceleration benefits are largely robust to the choice of decoding temperature, indicating that its performance gains stem from structural properties of the speculative mechanism rather than temperature-specific token distributions. This robustness makes RACER particularly attractive for real-world deployment, where decoding strategies may vary across applications without requiring retuning of acceleration hyperparameters.

\subsection{Parameter Robustness}
\label{sec:parameter_robustness}

\begin{figure*}[t]
    \includegraphics[height=0.22\textwidth]{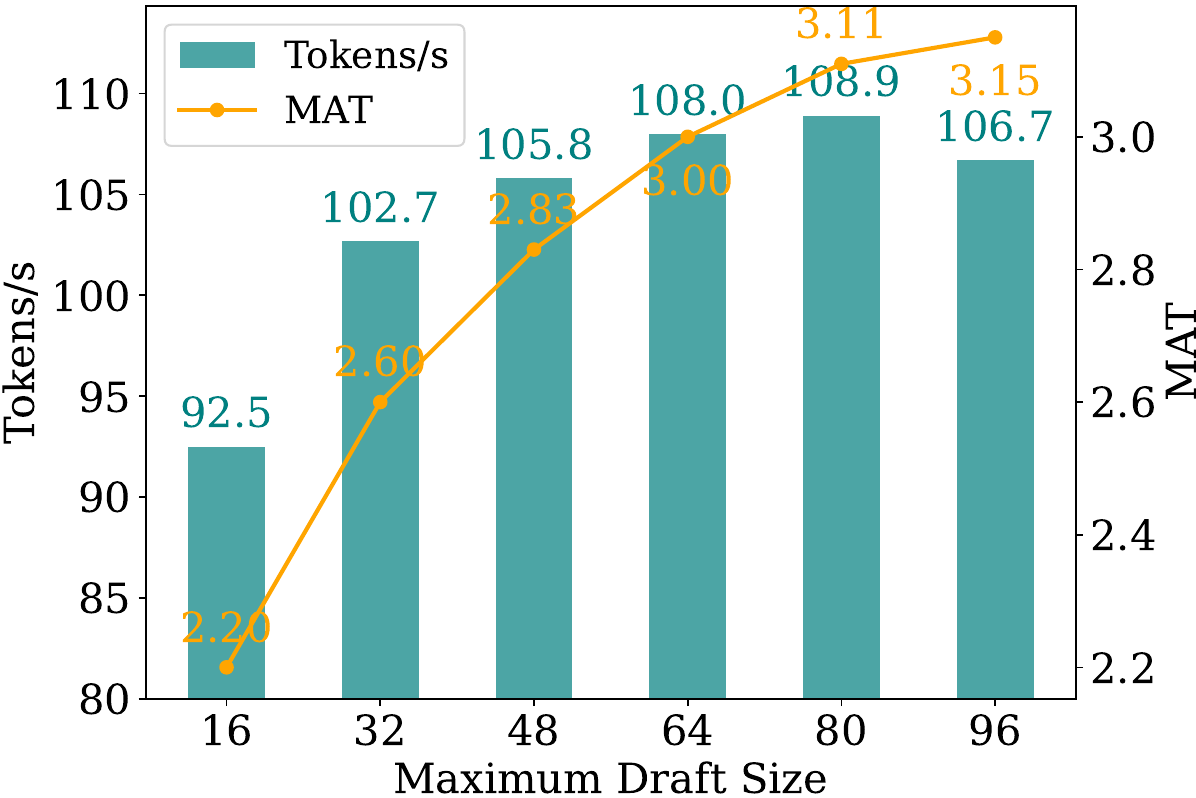} \hfill
    \includegraphics[height=0.22\textwidth]{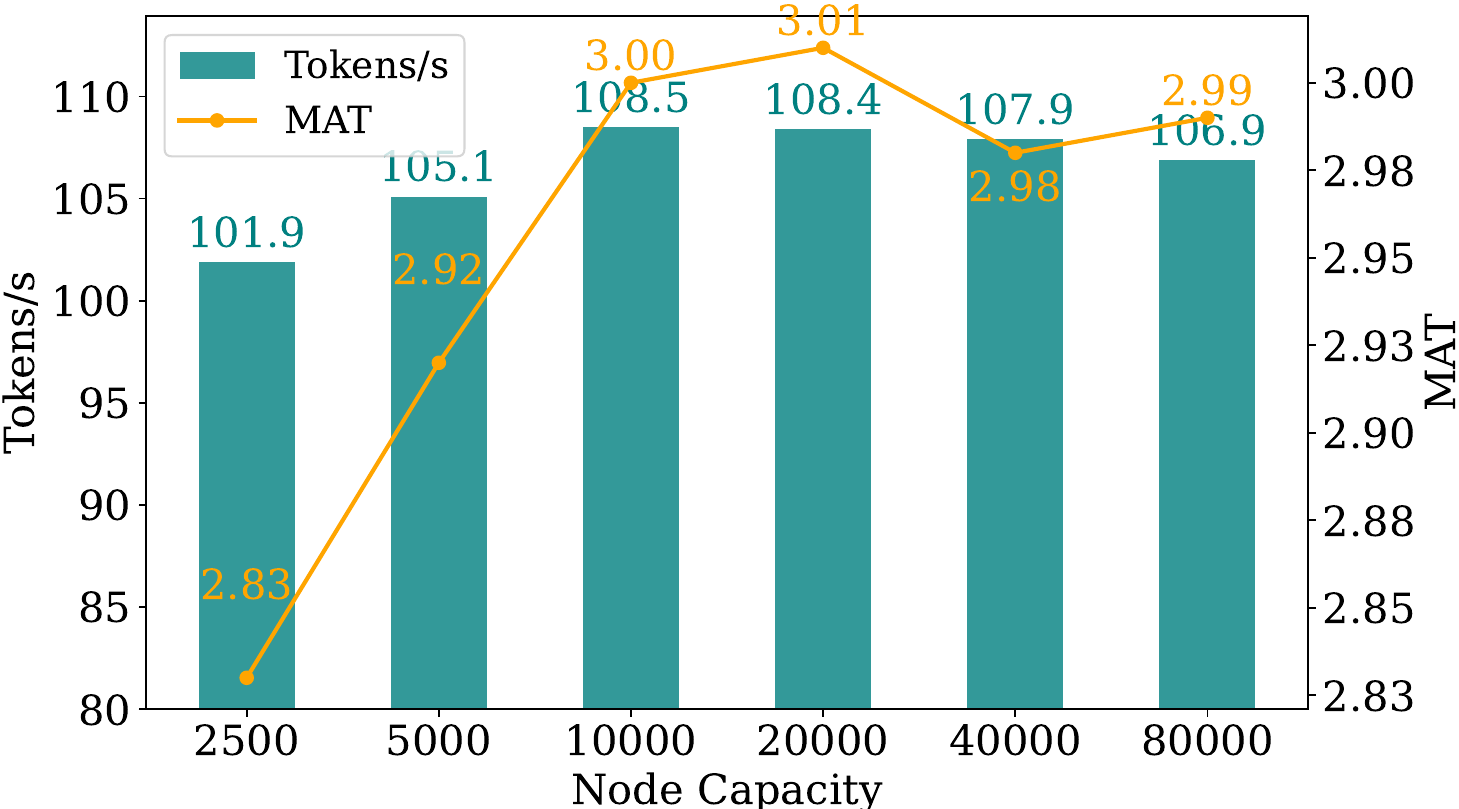} \hfill
    \includegraphics[height=0.22\textwidth]{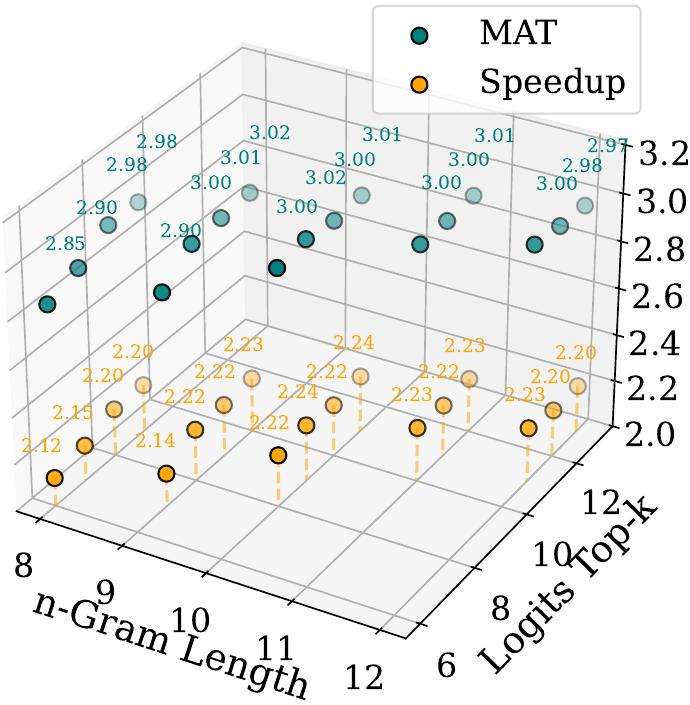} \hfill
    \caption{Ablation studies of RACER on key parameters: draft size, node capacity, $n$-gram depth (Retrieval Tree) and top-$k$ breadth (Logits Tree).}
    \label{fig:ablation}
\end{figure*}

We further study the robustness of RACER under different hyperparameter settings on Vicuna-7B-v1.5.
Conceptually, these hyperparameters control complementary aspects of the unified speculative draft:
the \emph{draft size} specifies the total number of draft tokens per step, the \emph{top-$k$ breadth} of the Logits Tree controls how widely we explore model-predicted candidates, and the \emph{capacity} and \emph{$n$-gram depth} of the Retrieval Tree determine how many $n$-grams can be stored and how long they can be.
A larger draft size or broader trees generally increases coverage and acceptance probability, but if they grow too large, the decoding regime may shift from memory-bound to compute-bound, leading to diminishing or even negative returns.
Similarly, increasing retrieval capacity and $n$-gram depth allows matching more rare patterns, but over-emphasizing long-tail matches can dilute the draft budget and reduce MAT.
In practice, RACER remains stable over a broad range of these settings.

Figure~\ref{fig:ablation} summarizes three ablation experiments.
\textbf{Draft Size (Figure a).}
The draft size controls the \emph{total} number of speculative tokens proposed by both the Logits Tree and the Retrieval Tree in each step.
Increasing the draft size from 16 to 64 steadily improves both MAT and throughput, after which the gains saturate.
This shows that RACER benefits from moderately larger drafts while remaining stable even with further expansion.
However, overly large drafts can push the system into a compute-bound regime, where the additional verification cost outweighs the benefit of higher acceptance.
The optimal draft size is therefore hardware-dependent: on resource-constrained or edge devices, it is often preferable to cap the draft size to match the device's most efficient batch inference mode.
\textbf{Retrieval Node Capacity (Figure b).}
The Retrieval Tree is implemented as an AC automaton whose \emph{capacity} specifies an upper bound on the number of $n$-gram states (e.g., 10K nodes).
Expanding the automaton's storage from 2.5K to around 10K-20K nodes yields the best trade-off between MAT and throughput.
Beyond this range, performance only fluctuates slightly, suggesting that RACER does not rely on excessively large retrieval buffers.
The built-in LRU eviction policy exploits both \emph{temporal} and \emph{spatial} locality: frequently reused $n$-grams are retained, while rarely used ones are pruned.
Since each node in an AC automaton has a unique parent and a failure link, the space complexity of the automaton $\mathcal{A}$ is $\mathcal{O}(|\mathcal{A}|)$, i.e., linear in the number of nodes.
In practice, this overhead is modest and remains negligible even with node sizes up to 10K.
\textbf{Joint Effect of $n$-gram Depth and Top-$k$ Breadth (Figure c).}
The $n$-gram depth controls the maximum height of the Retrieval Tree, and thus the longest context it can match.
With an upper bound such as $n=10$, the automaton can flexibly match keys of length $1$ to $9$ and retrieve the corresponding value sequences.
Because candidates are selected according to empirical frequency, this design naturally balances exploration and exploitation: it covers diverse patterns while prioritizing high-yield matches.
On the Logits Tree side, the top-$k$ breadth specifies how many high-probability continuations are expanded at each level; increasing it extends coverage into the long tail but also competes for the finite draft budget.
The 3D plot shows that both MAT and speedup improve smoothly as $n$-gram depth and top-$k$ breadth increase, and the performance surface remains relatively flat near the optimal range ($n$-gram depth 9-11, top-$k$ breadth 8-10), indicating that RACER is robust to small parameter deviations.
This trend is consistent with our analysis in Section~\ref{sec:logits_tree}, where the 85th percentile rank for the \emph{copy-logit} strategy is 9, suggesting that setting these parameters around this range is sufficient and provides a practical guideline when adapting RACER to new target models.

\section{More Implementation Details}
\label{sec:impl_detail}

\subsection{Logits Tree Construction}

\begin{algorithm*}
\caption{Construct Logits Tree (BFS) and Return Draft Candidates}
\label{alg:logits_tree_bfs}
\begin{algorithmic}[1]
\Function{BuildLogitsTree}{next\_token, $C$} \Comment{$C$ is the maximum number of draft nodes}
    \State Initialize an empty queue $Q$
    \State Initialize an empty list candidates
    \State token(0) $\gets$ next\_token
    \State \textbf{push} $Q \gets$ 0
    \State $C \gets C - 1$  \Comment{Root consumes one draft slot}
    \While{not $Q$.isEmpty()}
        \State $u \gets Q$.dequeue()
        \If{$C>0$}
            \State next\_breadth $ \gets \begin{cases}
                b($u$), & \text{if } u=\text{root} \\ 
                \lfloor b($u$)/2 \rfloor, & \text{otherwise}
            \end{cases}$
            \Comment{Breadth allocation follows Eq.~\ref{eq:breadth}}
            \For{$j \gets 0$ \textbf{to} $b$($u$) - 1}
                \If{$C=0$} \State \textbf{break} \EndIf
                \State $v \gets k\times u + j$ \Comment{The $j$-th child of $u$}
                \State token($v$) $\gets$ top\_k[token($u$)][j]
                \State $b(v) \gets$ next\_breadth
                \State \textbf{push} $Q \gets v$
                \State next\_breadth $\gets \lfloor \text{next\_breadth}/2 \rfloor$ \Comment{Breadth allocation follows Eq.~\ref{eq:breadth}}
                \State $C \gets C - 1$
            \EndFor
        \Else \Comment{Backtrack to get a draft candidate}
            \State path $\gets [\ ]$; \ $v \gets u$
            \While{$v \neq \epsilon$}
                \State \textbf{append} token($v$) to path
                \State $v \gets$ parent($v$)
            \EndWhile
            \State \textbf{reverse}(path) \Comment{Leaf$\to$root to root$\to$leaf}
            \State \textbf{append} path to candidates
        \EndIf
    \EndWhile
    \State \Return candidates
\EndFunction
\end{algorithmic}
\end{algorithm*}

\begin{figure*}[htbp]
    \centering
    \includegraphics[width=1\linewidth]{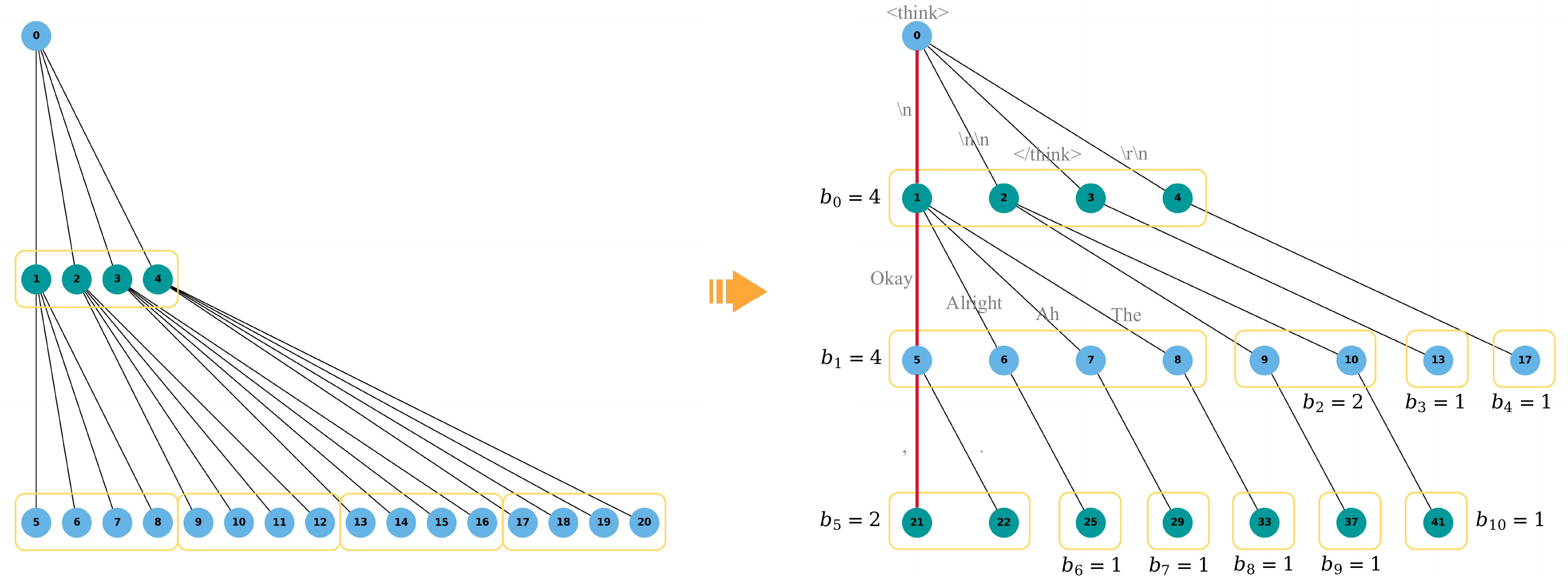}
    \caption{The tree on the left shows the expansion of an unpruned 4-ary tree with 21 nodes, while the tree on the right depicts the expansion of the pruned 4-ary tree with the same number of nodes.
    The path in red represent a possible candidate \texttt{[<think>, <end\_of\_line>, Okay, <comma>]}.}
    \label{fig:logits_tree}
\end{figure*}

The Logits Tree leverages the top-$k$ adjacency matrix from past logits and expands breadth-first, following the breadth allocation rule in Eq.~\ref{eq:breadth}. The full procedure is described in Algorithm~\ref{alg:logits_tree_bfs}. Figure~\ref{fig:logits_tree} illustrates how Logits Tree is pruned from dense to sparse, with the same capacity of 21.

\subsection{Aho--Corasick Automaton Construction and Transition} 
\label{sec:ac_automata}

The Aho--Corasick automaton could be simply described as trie with failure links. Each failure link at a node connects to the longest proper suffix of the string at that node, which also serves as a prefix for another pattern in the trie. If no such suffix exists, the link reverts to the root. This is analogous to the ``failure function'' in the Knuth--Morris--Pratt (KMP) string-matching algorithm~\citep{knuth1977fast}, but Aho--Corasick extends this idea to work efficiently for multiple patterns. 

\begin{figure*}[htbp]
    \centering
    \includegraphics[width=1\linewidth]{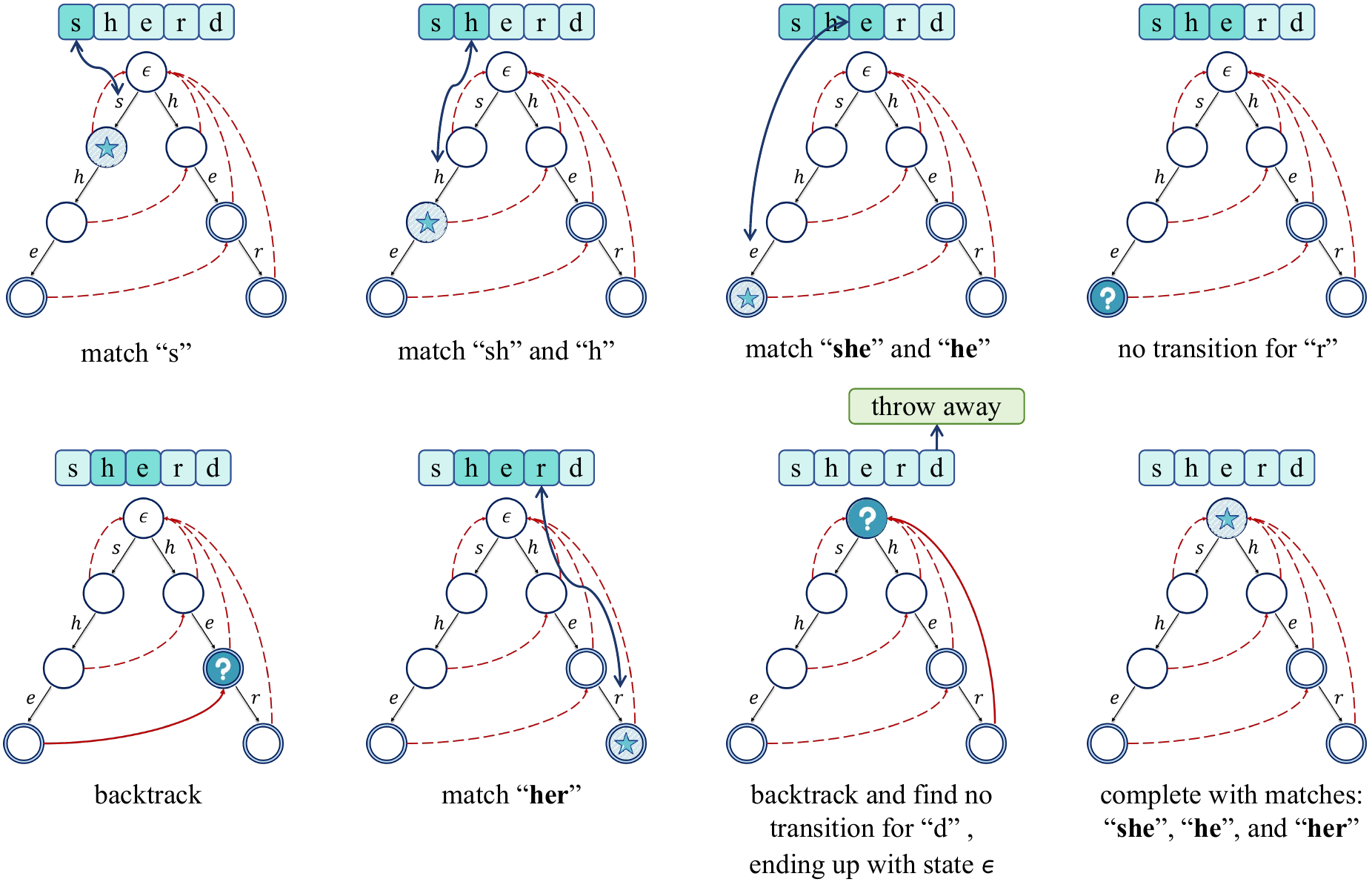}
    \caption{The process of how ``sherd'' matches patterns ``she'', ``he'', and ``her'' by transitions on an AC automaton.}
    \label{fig:ac_trans}
\end{figure*}

\begin{figure}
    \centering
    \includegraphics[width=0.8\columnwidth]{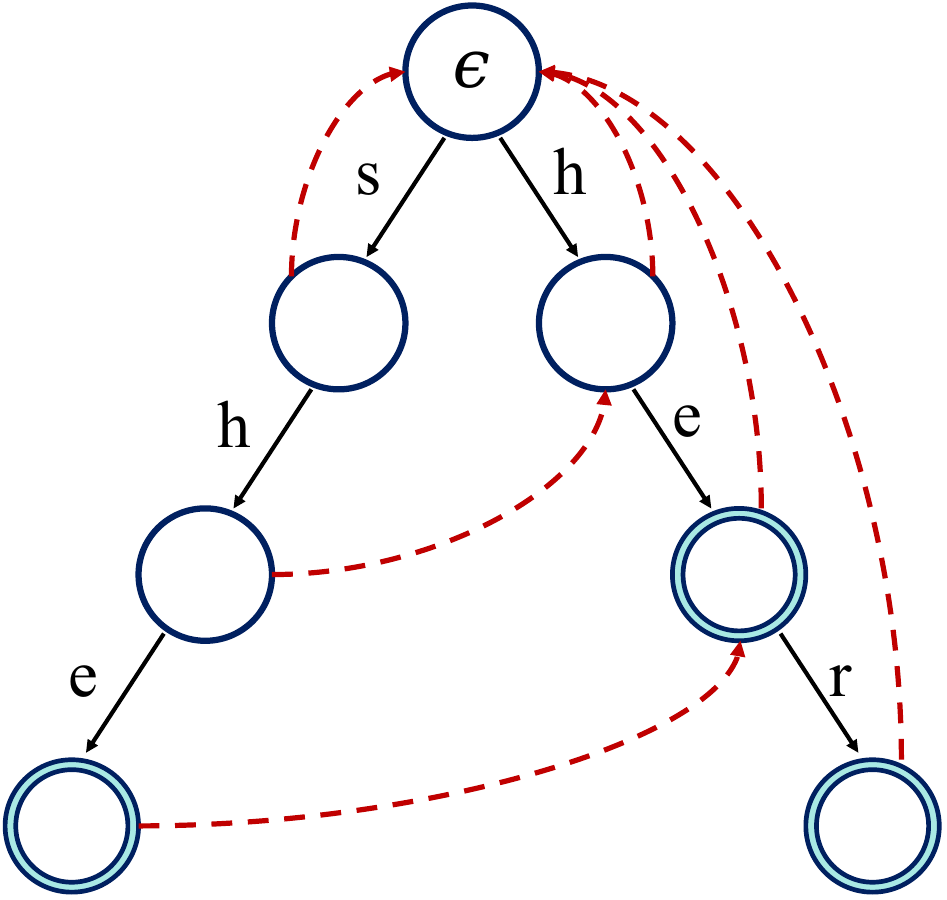}
    \caption{The illustration of an Aho--Corasick automaton with patterns ``she'', ``he'', and ``her'', with final states in double circles and failure links in red.
    }
    \label{fig:automaton_orig}
\end{figure}

Figure \ref{fig:automaton_orig} illustrates the AC automaton's structure, showcasing failure links in red and final states with double circles, though some transitions may be omitted for clarity. The process begins with an input sequence that progresses through the trie. If a mismatch occurs, such as when at the state ``she'' and the next input is ``r'' without a corresponding edge, the automaton utilizes failure links to backtrack until a valid node with the ``r'' edge is found or until it returns to the root. When the automaton reaches the state ``her'', not only is the pattern ``her'' itself recognized but the state of its failure pointer is also included. This forms part of a recursive process: matching a state involves sequentially matching the state of its failure pointer until it traces back to the root node, which represents the absence of further matches, denoted as $\epsilon$.

For further clarification, Figure~\ref{fig:ac_trans} illustrates the matching process for the string ``sherd'', identifying the substrings ``she'', ``he'', and ``her''. The elements highlighted in dark blue represent both the longest pattern prefix that the current state can match, and the minimal suffix information necessary for subsequent matches. This setup can be visualized as a ``sliding window'' that moves from left to right across each position. During normal transitions, this sliding window accordingly steps to the right. Conversely, during backtracking, the state transitions via the failure pointer, effectively discarding any irrelevant left-side components. Note that in actual implementations, the failure links in AC automata are primarily used during the construction phase and the match phase. Once the automaton is constructed, these failure links are often replaced by virtual transitions that directly lead to the correct states like Algorithm~\ref{alg:trans}. This optimization streamlines the matching process, enhancing efficiency by reducing unnecessary transitions.

\begin{algorithm*}
\caption{Calculate failure links for nodes in an AC automaton}
\label{alg:trans}
\begin{algorithmic}[1]
\Function{get\_transitions}{}
    \State Initialize an empty queue $Q$
    \For{$v \gets$ children of the root}
        \State fail($v$) $\gets$ root \Comment{Set initial failure state to  root}
        \State Enqueue $v$ into $Q$
    \EndFor
    \While{not $Q$.isEmpty()}
        \State $u \gets Q$.dequeue()
        \For{$i \gets$ possible transitions from $u$}
            \State $v \gets$ child($u$, $i$)
            \If{$v \neq \epsilon$}
                \State fail($v$) $\gets$ child(fail($u$), $i$) \Comment{Update the failure pointer}
                \State Enqueue $v$ into $Q$
            \Else
                \State $f \gets$ fail($u$)
                \While{$f \neq$ root \textbf{and} child($f$, $i$) $= \epsilon$}
                    \State $f \gets$ fail($f$)
                    \Comment{Backtrack through the failure pointer}
                \EndWhile
                \If{child($f$, $i$) $\neq \epsilon$}
                    \State fail($v$) $\gets$ child($f$, $i$)
                \Else
                    \State fail($v$) $\gets$ root
                \EndIf
            \EndIf
        \EndFor
    \EndWhile
\EndFunction
\end{algorithmic}
\end{algorithm*}

\subsection{LRU Eviction Strategy}

\begin{algorithm*}
\caption{LRU Eviction Strategy in AC Automaton}
\label{alg:lru_eviction}
\begin{algorithmic}[1]
\Function{Touch}{node}
    \State Move node to the front of LRU\_LIST
    \State Update LRU\_MAP[node] $\gets$ LRU\_LIST.begin()
\EndFunction

\Function{TouchPrefix}{node}
    \While{node $\neq \epsilon$}
        \State fail(node) $\gets$ root \Comment{Default failure link to the root before rebuild}
        \State \Call{Touch}{node}
        \State node $\gets$ parent(node)
    \EndWhile
\EndFunction

\Function{TransTokens}{tokens}
    \State $u \gets$ curren\_state
    \For{each $t \in$ tokens}
        \State \Call{Touch}{$u$}
        \While{$u \neq$ root \textbf{and} child($u, t$) $= \epsilon$}
            \State $u \gets$ fail($u$) \Comment{Failure transition}
        \EndWhile
        \State \Call{TouchPrefix}{$u$} \Comment{Update prefix after failure transition}
        \If{child($u, t$) $\neq \epsilon$}
            \State $u \gets$ child($u, t$)
        \Else
            \State $u \gets$ root
        \EndIf
    \EndFor
    \State \Call{Touch}{$u$} \Comment{Final state after processing tokens}
    \State current\_state $\gets u$
\EndFunction

\Function{InsertTokens}{tokens, frequency}
    \State $u \gets$ root
    \State freq($u$) $\gets$ freq($u$) + frequency
    \For{each $t \in$ tokens}
        \State \Call{Touch}{$u$}
        \If{child($u, t$) $= \epsilon$}
            \State new\_node $\gets$ LRU\_LIST.back()
            \State new\_node.reset()
            \State child($u, t$) $\gets$ new\_node
        \EndIf
        \State $u \gets$ child($u, t$)
        \State freq($u$) $\gets$ freq($u$) + frequency
    \EndFor
    \State \Call{Touch}{$u$} \Comment{Touch the leaf node}
\EndFunction
\end{algorithmic}
\end{algorithm*}

\begin{figure*}[t]
    \centering
    \includegraphics[width=0.32\textwidth]{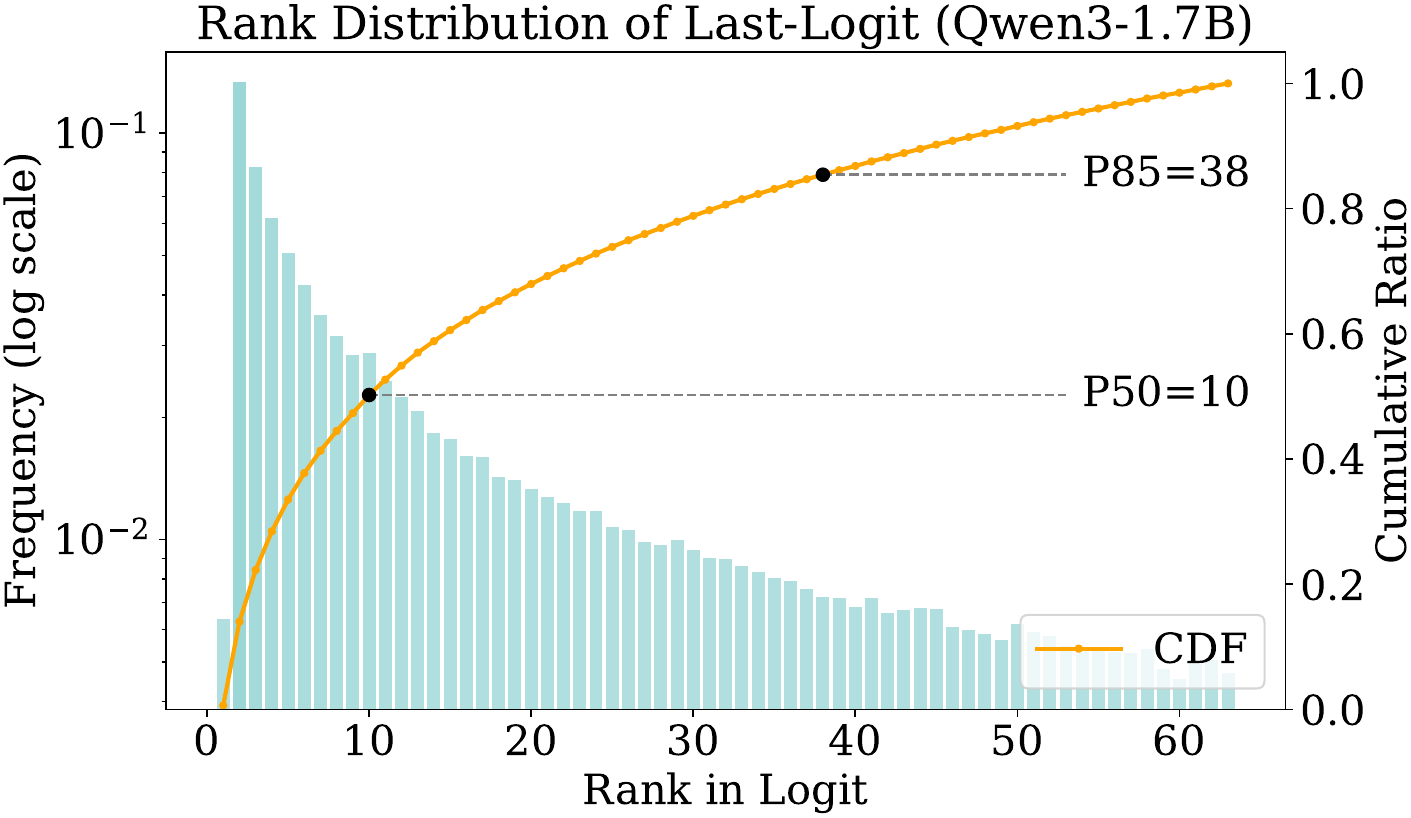} \hfill
    \includegraphics[width=0.32\textwidth]{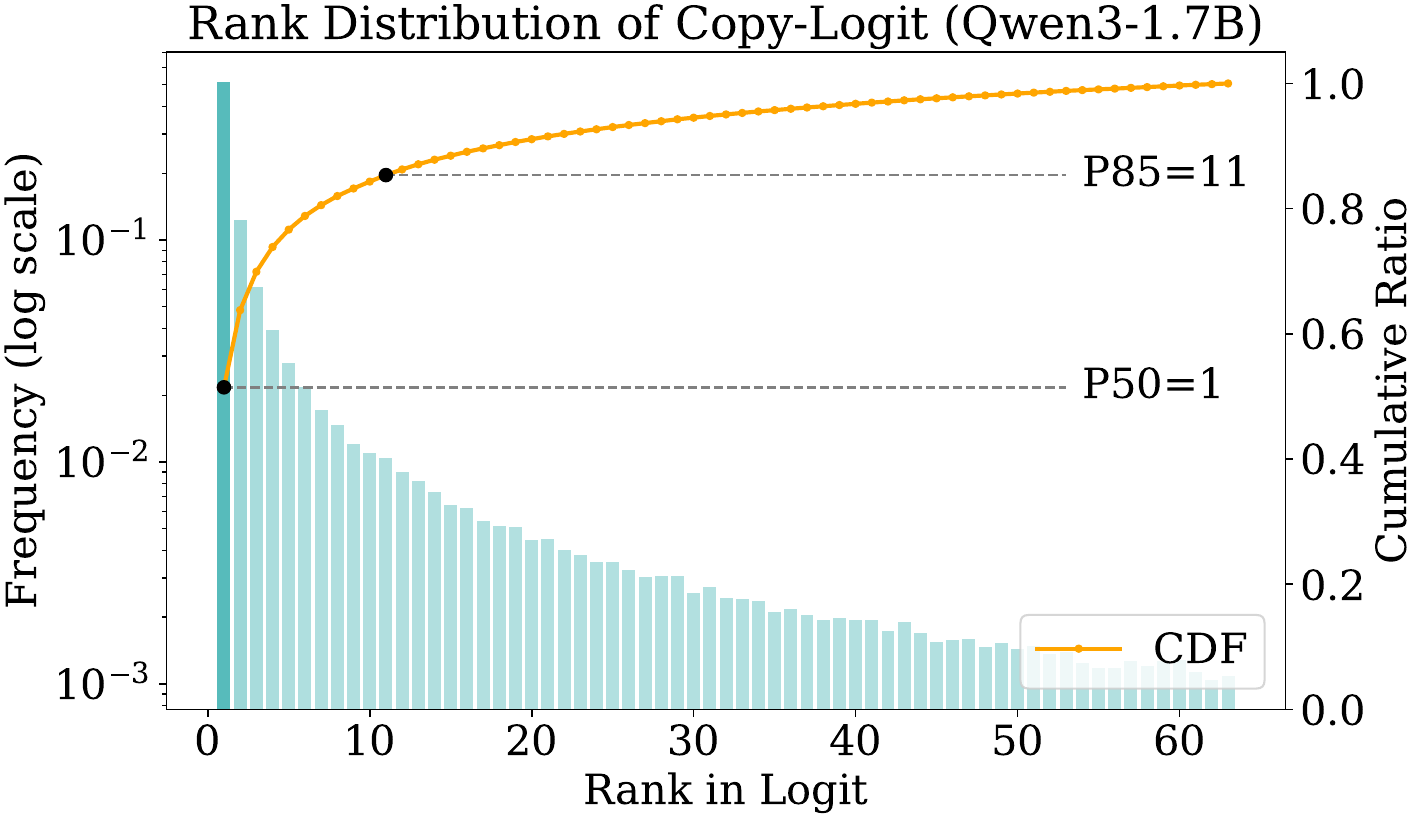} \hfill
    \includegraphics[width=0.32\textwidth]{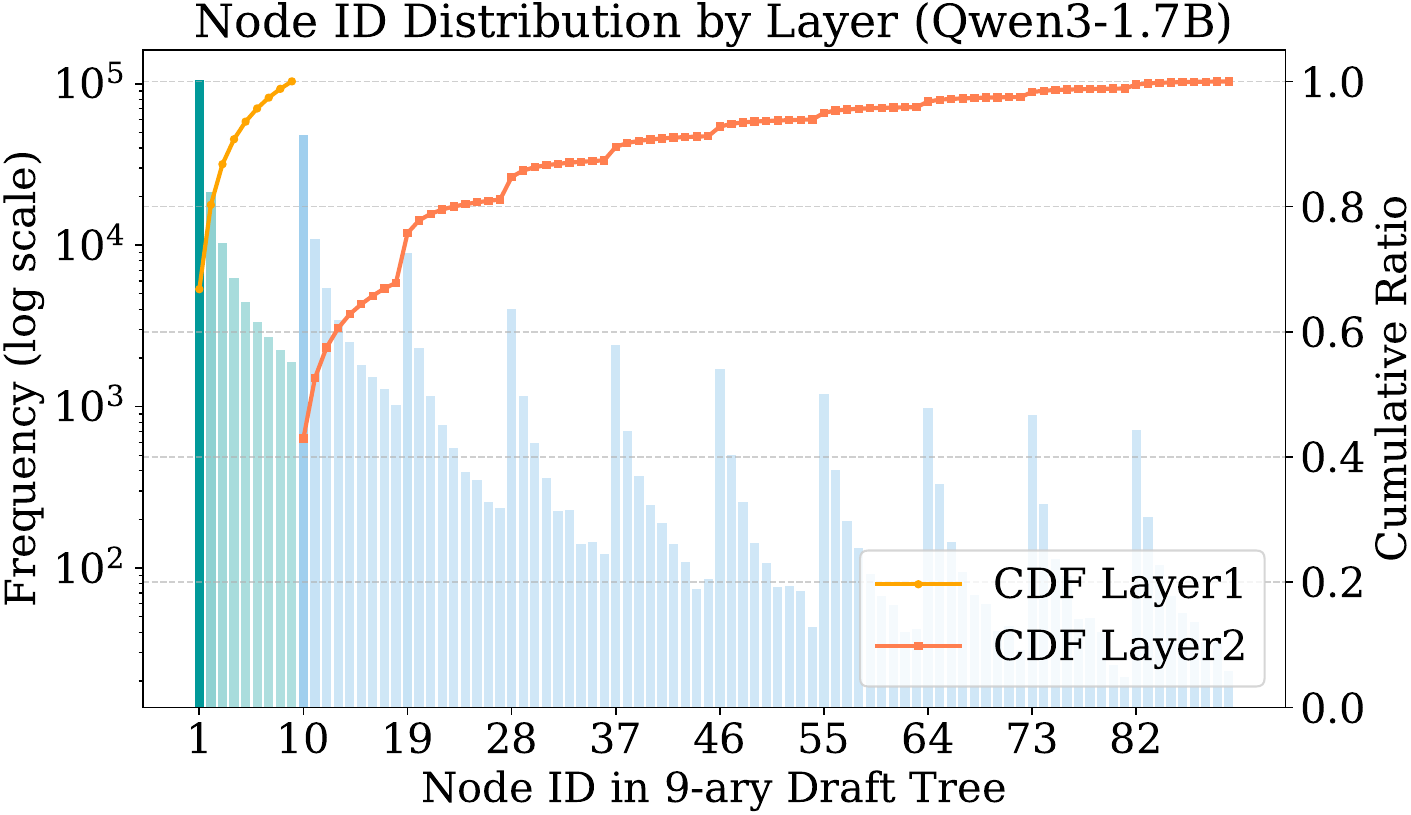} \\
    \includegraphics[width=0.32\textwidth]{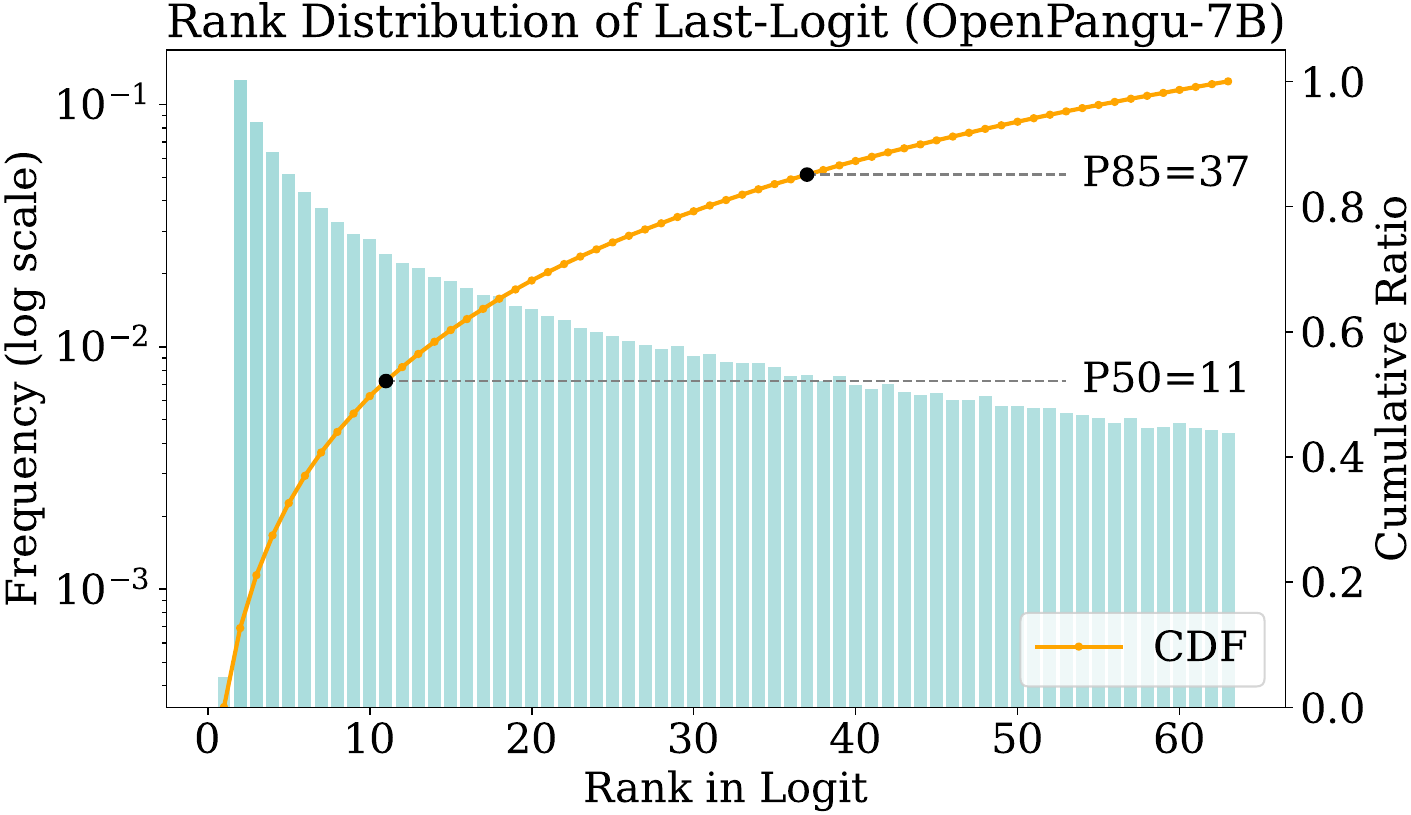} \hfill
    \includegraphics[width=0.32\textwidth]{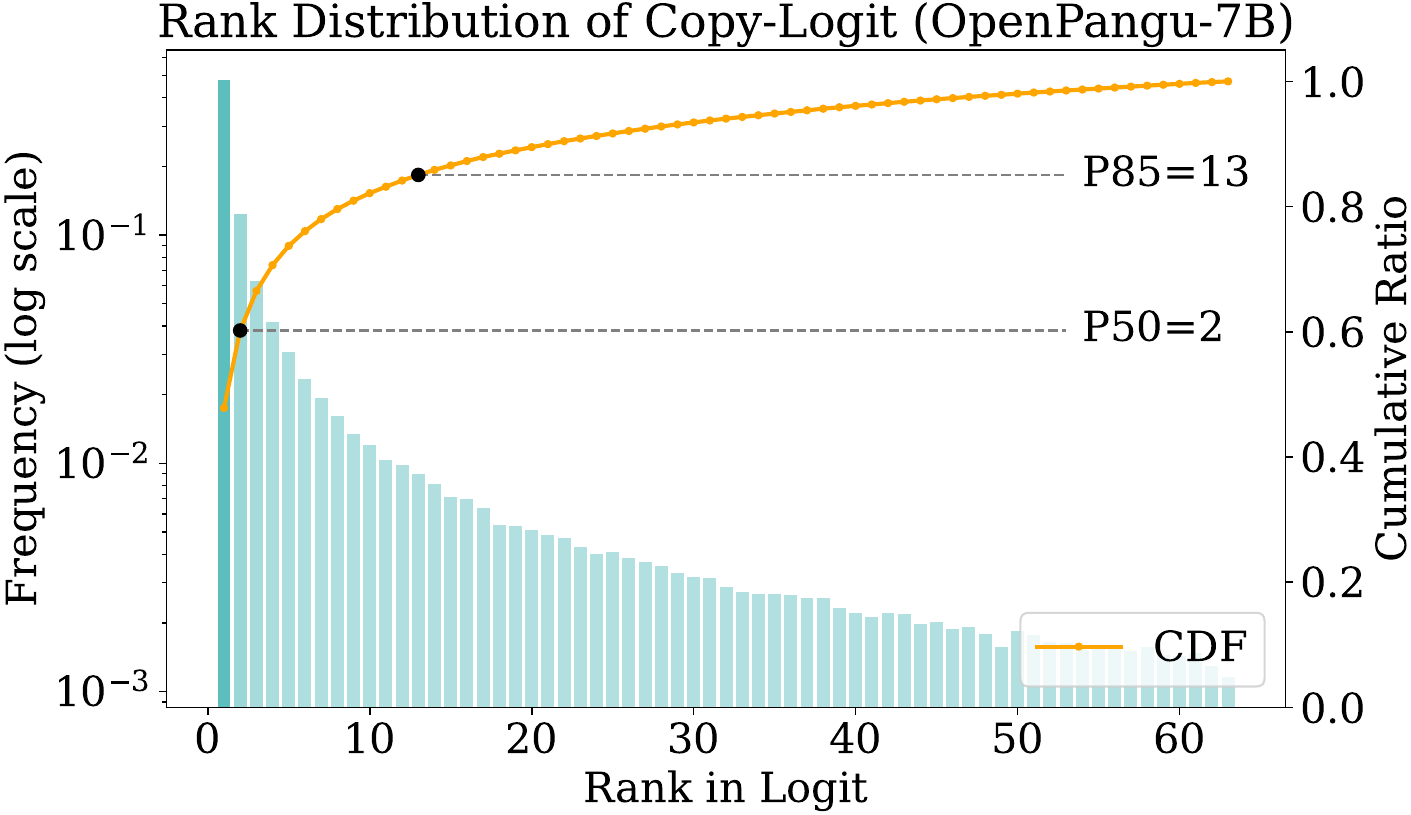} \hfill
    \includegraphics[width=0.32\textwidth]{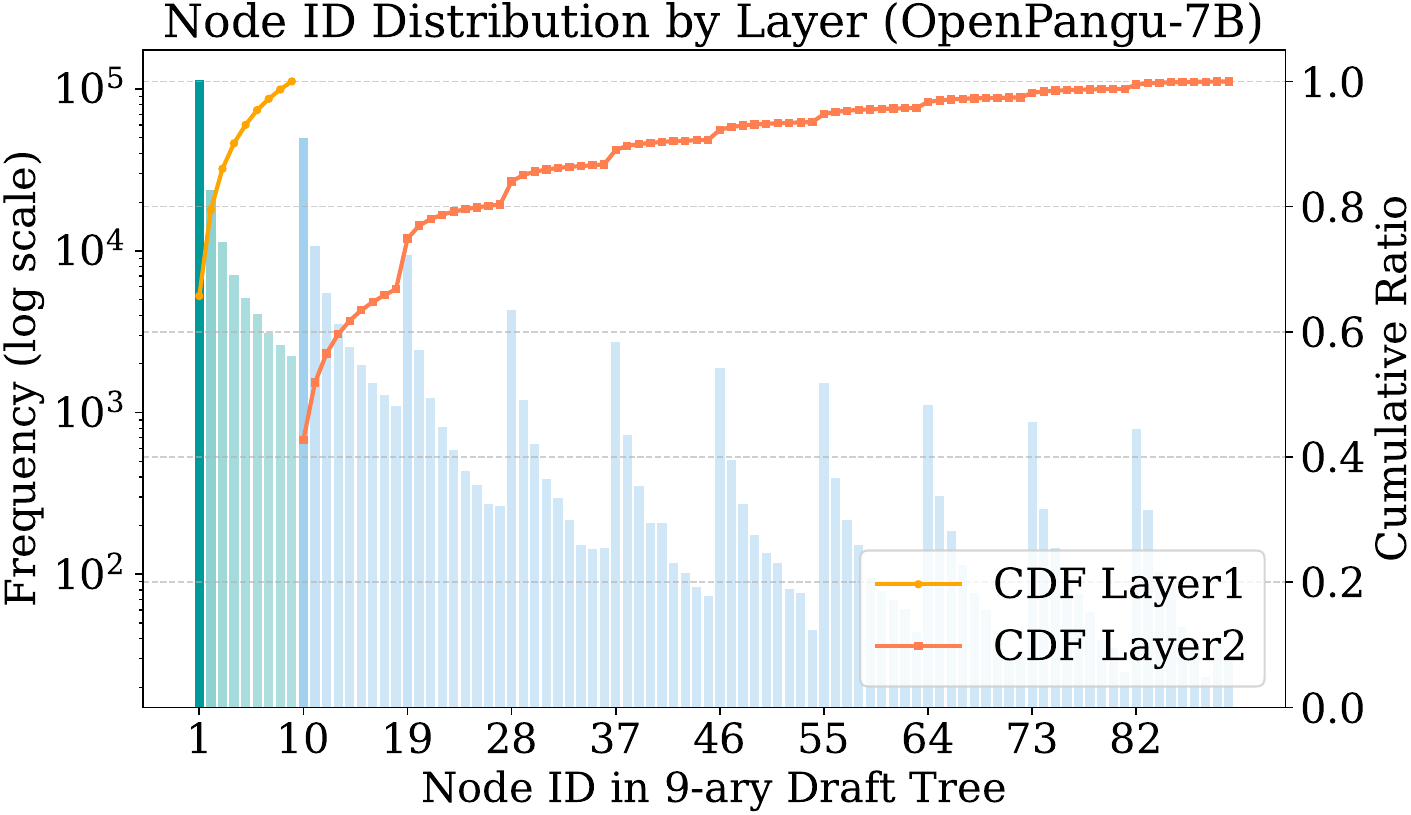} 
    \caption{Accepted draft statistics of Qwen3-1.7B and OpenPangu-7B on Spec-Bench: \textit{(Left)} Logits Tree with one layer beyond next-token expanded with \textit{last-logit}. \textit{(Middle)} Logits Tree with one layer beyond next-token expanded with \textit{copy-logit}. \textit{(Right)} Logits Tree with two layers beyond next-token expanded with \textit{copy-logit} in $9$-ary manner.}
    \label{fig:sup_logits_tree}
\end{figure*}

In RACER's retrieval automaton, we adopt an LRU (Least Recently Used) eviction mechanism to manage limited node capacity. When the maximum number of nodes is reached, the least recently accessed node is recycled and reassigned to represent a new state. This ensures that the automaton continuously adapts to the most relevant $n$-grams from the current decoding context while maintaining bounded memory.

The mechanism is shown in Algorithm~\ref{alg:lru_eviction} and works as follows: \textbf{Touch}: Every time a node is visited, it is moved to the front of the LRU list and its reference is updated in the hash map. This guarantees that the tail of the list always contains the least recently used node. \textbf{TouchPrefix}: When a failure transition occurs, not only the current node but also its ancestors along the prefix are ``touched''. This ensures that the entire matching path is marked as recently used. \textbf{TransTokens}: When processing a sequence of tokens, the automaton repeatedly performs \emph{Touch} and failure transitions until either a matching child node is found or the traversal falls back to the root. \textbf{InsertTokens}: When inserting a new $n$-gram, if no free node is available, the node at the back of the LRU list (the least recently used one) is evicted and reset. It is then reassigned as the child for the new transition. Frequency counters along the insertion path are updated accordingly.
This design ensures that: Only leaf nodes are evicted, preventing structural corruption of the automaton. The automaton remains adaptive to changing contexts, exploiting temporal and spatial locality during decoding. The eviction and update operations remain lightweight and efficient, keeping inference fast.

\paragraph{Time Complexity}
The Touch operation runs in constant time $\mathcal{O}(1)$, as it only updates the doubly linked list and hash map. TouchPrefix has a worst-case complexity of $O(d)$, where $d$ is the maximum depth of the automaton (i.e., the $n$-gram length, typically a small constant). The TransTokens procedure processes each token in the input sequence and may backtrack up to depth $d$ through failure links, giving a worst-case complexity of $\mathcal{O}(|\text{tokens}| \cdot d)$, though in practice it is close to linear $O(|\text{tokens}|)$. Finally, InsertTokens requires at most $d$ steps for each token sequence, resulting in $O(|\text{tokens}|)$ complexity.
\paragraph{Space Complexity}
The storage requirement is linear in the number of automaton nodes. Trie nodes occupy $O(|\mathcal{A}|)$ space, bounded by the maximum capacity of the automaton. The LRU list and hash map also require $O(|\mathcal{A}|)$ space, as each node is tracked in both structures. Thus, total memory is $O(|\mathcal{A}|)$, capped in practice at $10^4$-$10^5$ nodes, which corresponds to tens of megabytes -- well within the budget of modern devices and suitable for memory-constrained scenarios.

\subsection{Case Study of Retrieval Expansion}
\label{sec:case_study_expansion}

Suppose the current match state is \texttt{[<think>, </think>]}, and the automaton has observed:
\begin{itemize}
\item \texttt{[<think>, </think>, Okay]} with frequency $3$,
\item \texttt{[</think>, Yes, <space>]} with frequency $2$,
\item \texttt{[</think>, Yes, <comma>]} with frequency $1$.
\end{itemize}
Pooling continuations over all prefixes ending in \texttt{</think>} yields:
$$
\begin{aligned}
& \texttt{[Okay]}: 3,\quad
  \texttt{[Yes]}: 2{+}1=3,\\
& \texttt{[Yes, <space>]}: 2,\quad
  \texttt{[Yes, <comma>]}: 1.
\end{aligned}
$$

Selecting the top-$3$ continuations gives:
$$
\begin{aligned}
& \texttt{[Okay]}: 3,\quad
  \texttt{[Yes]}: 2{+}1=3,\\
& \texttt{[Yes, <space>]}: 2,\quad
\end{aligned}
$$

If a depth constraint is applied (i.e., only continuations consistent with the matched suffix of length $2$ are allowed), then only the continuation \texttt{Okay} remains valid, because it is the only 3-gram whose prefix exactly matches the current border \texttt{[<think>, </think>]}.  
Other candidates such as \texttt{[Yes]} or \texttt{[Yes, <space>]} originate from shorter matches ending in \texttt{</think>}, and are therefore pruned under the depth restriction.

\section{Futher Explanation of Logits Tree}
\label{sec:explanation_logits_tree}

The tokenizer, with vocabulary $\mathcal{V}$, maps the text into discrete tokens. At each decoding step, one token is determined at a time, resulting in a token sequence that represents a specific forward path. For a path with $k$ tokens, the LLM narrows the search space $\mathcal{V}^k$ using either greedy or nucleus sampling strategies. Speculative decoding with a draft tree can be viewed as a search with a scope broader than that of the LLM, but narrower than the entire exponential space.

It is reasonable to model the \textit{seen information} using frequency as an indicator of the confidence in the likelihood of a continuation occurring in the future, given the long-tail property of natural language. Thus, we select $n$-grams for the Retrieval Tree. However, for the unseen portion, the hidden states of LLMs encode more than just the next token~\citep{mehra2025multi, dong2025emergent}. This implies that the top-$k$ logits not only reflect the next token but also include some subsequent tokens. As a limiting case, setting $k$ to the vocabulary size $|\mathcal{V}|$, which can range up to 32K or even 150K, would result in a \textit{copy-logit} accuracy of $100\%$. However, due to the long-tail property of language, it is clear that covering the entire vocabulary is unnecessary. A significantly smaller $k$ (e.g., 63, as used in our preliminary experiment) can still capture the advantages from the top spikes in the probability distribution. To further investigate this, we conduct ablation studies on the breadth and examine how different logit-reuse strategies influences the mean accepted tokens (MAT).

\begin{figure}[htbp]
    \centering
    \includegraphics[width=0.8\columnwidth]{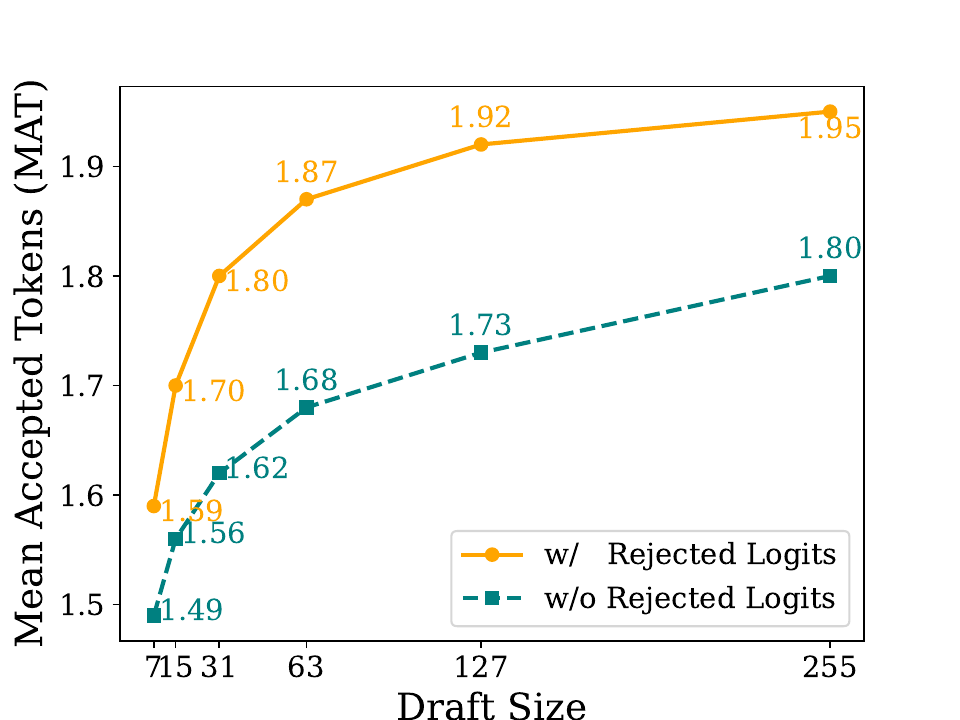}
    \caption{
        Ablation study on search breadth (top-$k$) for extending only one layer with \textit{copy-logit} using Vicuna-7B. 
    }
    \label{fig:ablation_copylogit}
\end{figure}

Two settings will be discussed. \textbf{w/ Rejected Logits:} Reuse all $k$ logits, even those rejected by the current verification. \textbf{w/o Rejected Logits:} Reuse at most two logits from the next token and one draft token (only if accepted).
Both settings have the same limit when $k = |\mathcal{V}|$, where the MAT will be 2 (one sampled next token and one accepted draft token). Reusing rejected logits could provide updated context information for future reuse. For example, if AB and AC are proposed with prefix X, and only AB is accepted, the rejected logit of AC could still be used to approximate the logit of ABC. As a result, the first setting could accumulate advantages, narrowing the search scope more effectively than the second one.

Figure~\ref{fig:ablation_copylogit} demonstrates that reusing rejected logits consistently results in more accepted tokens, even when only 7 draft tokens are used (1.59 vs. 1.49). As the draft size increases, the gap between the two settings also widens. When using a draft size of 255 logits, the MAT reaches 1.95, which is very close to the theoretical maximum of 2. In contrast, reusing only 2 logits results in a MAT of 1.80. This suggests that the accumulative pruning of the draft scope is achieved with the first setting, and this attribute is crucial in the self-speculative procedure.

\begin{figure}[htbp]
    \centering
    \includegraphics[width=0.8\columnwidth]{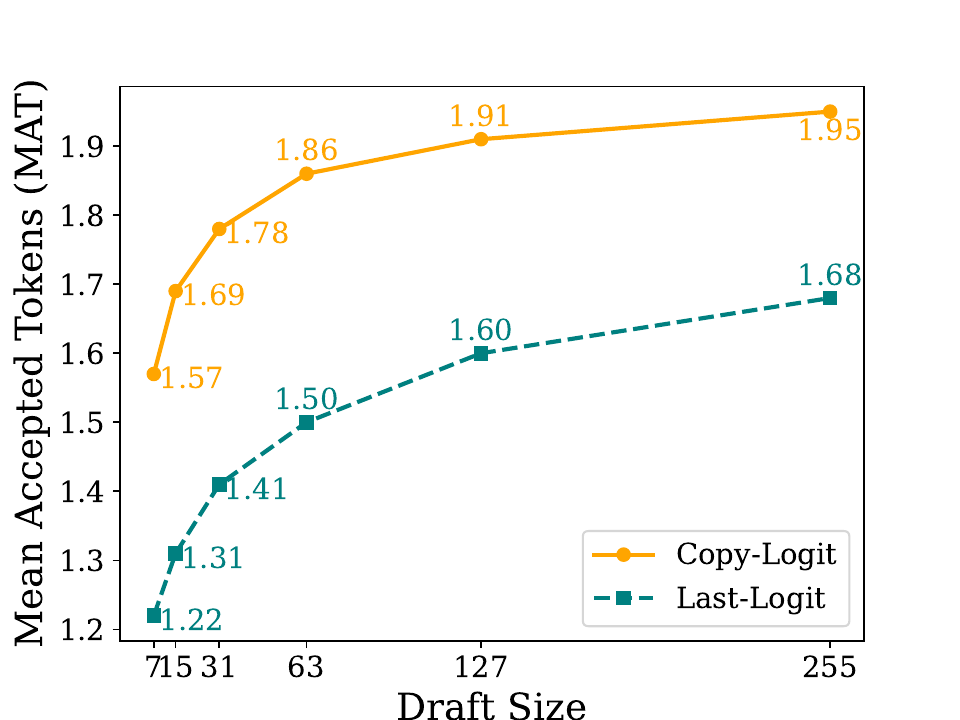}
    \caption{
        Ablation study on search breadth (top-$k$) for extending only one layer with \textit{copy-logit} and \textit{last-logit} using Qwen3-8B.
    }
    \label{fig:ablation_logit}
\end{figure}

In Figure~\ref{fig:ablation_logit}, \textit{copy-logit} continues to improve MAT as $k$ increases, approaching 2.0 when $k=127/255$, whereas \textit{last-logit} exhibits significantly slower growth and remains at 1.68 even at $k=255$. This confirms that the heavy-tail advantage of copy-logit is not restricted to a single model or a single  setting.

Supporting this observation, Figure~\ref{fig:sup_logits_tree} and the supplementary experiments confirm the same trend seen in the main results (Figure~\ref{fig:first_layer} and Figure~\ref{fig:kary_tree}). Specifically, the \emph{copy-logit} expansion consistently produces a sharper, heavy-tailed acceptance distribution compared to \emph{last-logit}, with most accepted tokens concentrated in the top ranks. Furthermore, deeper $k$-ary expansions preserve this concentration and further validate the breadth allocation rule. These findings confirm that the phenomena discussed in the main text are robust across different model scales and settings, thereby reinforcing our choice of \textit{copy-logit} as the default expansion strategy in RACER.

Actually, the expansion of the Logits Tree can be interpreted as a form of
``beam search'' in which the exact scores of individual paths are unavailable
and are instead approximated via head dominance.
Referring to Eq.~\ref{eq:breadth}, let the root breadth be $b_0 = k$, and define
$m = \lfloor \log_2(k) \rfloor$.
By binary decomposition, $k$ can be written as
\begin{equation}
\begin{aligned}
k
&= \sum_{i = 0}^{m} \alpha_i 2^{i},
& \alpha_i \in \{0,1\}, \ \alpha_m = 1, \\
&= 2^{m} + r,
& r = \sum_{i = 0}^{m-1} \alpha_i 2^{i}, \ 0 \le r < 2^{m}.
\end{aligned}
\end{equation}

Under this formulation, the maximum number of nodes in any single layer
(i.e., the maximum layer breadth) can be expressed as
\begin{equation}
\label{eq:max_breadth}
\begin{aligned}
\mathcal{L}(k) =
& \sum_{i = 0}^{m - 1} \alpha_i (i+2) 2^i
+ 2^{m}\cdot m + 1.
\end{aligned}
\end{equation}

We show that $\mathcal{L}(k)$ is upper bounded by
$k\lceil \log_2(k) \rceil + 1$, and that the first occurrence of the maximum
layer appears at depth $\lfloor \log_2(k) \rfloor + 1$.

\begin{table*}[htbp]
\centering
\caption{Results on reasoning tasks evaluated on NVIDIA A800 (80GB): GSM8K, AIME, and MATH, 
reported in mean accepted tokens (MAT) and speedup ratio. $\dagger$ denotes that the EAGLE-3 model weight is from AngelSlim's re-implementation.}
\small
\label{tab:mix_a800_eagle_reasoning}
\begin{threeparttable}
\renewcommand{\arraystretch}{1.2}
\setlength{\tabcolsep}{5pt}
\begin{tabular}{l l cc cc cc cc}
\toprule
\textbf{Model} & \textbf{Method} 
& \multicolumn{2}{c}{\textbf{GSM8K}} 
& \multicolumn{2}{c}{\textbf{AIME}} 
& \multicolumn{2}{c}{\textbf{MATH}} 
& \multicolumn{2}{c}{\textbf{Average}} \\
\cmidrule(lr){3-4} \cmidrule(lr){5-6} \cmidrule(lr){7-8} \cmidrule(lr){9-10}
& & MAT & Speedup & MAT & Speedup & MAT & Speedup & MAT & Speedup \\
\midrule
\multirow{2}{*}{Qwen3 8B}
 & EAGLE-3$^\dagger$ & \textbf{3.86} & 2.65 & \textbf{3.44} & 2.44 & \textbf{3.55} & \textbf{2.60} & 3.62 & 2.56 \\
 & \cellcolor{teal!20}\textbf{RACER} & \cellcolor{teal!20}3.01 & \cellcolor{teal!20}\textbf{2.68} & \cellcolor{teal!20}2.91 & \cellcolor{teal!20}\textbf{2.63} & \cellcolor{teal!20}2.88 & \cellcolor{teal!20}2.58 & \cellcolor{teal!20}2.93 & \cellcolor{teal!20}\textbf{2.63} \\
\midrule
\multirow{2}{*}{Qwen3 14B}
 & EAGLE-3$^\dagger$ & \textbf{3.08} & 2.23 & \textbf{3.05} & 2.24 & \textbf{3.06} & 2.23 & \textbf{3.06} & 2.23 \\
 & \cellcolor{teal!20}\textbf{RACER} & \cellcolor{teal!20}2.95 & \cellcolor{teal!20}\textbf{2.72} & \cellcolor{teal!20}2.90 & \cellcolor{teal!20}\textbf{2.64} & \cellcolor{teal!20}2.87 & \cellcolor{teal!20}\textbf{2.55} & \cellcolor{teal!20}2.91 & \cellcolor{teal!20}\textbf{2.64} \\
\midrule
\multirow{2}{*}{Qwen3 32B}
 & EAGLE-3$^\dagger$ & \textbf{3.32} & 2.51 & \textbf{3.26} & \textbf{2.34} & \textbf{3.33} & \textbf{2.42} & \textbf{3.30} & \textbf{2.42} \\
 & \cellcolor{teal!20}\textbf{RACER} & \cellcolor{teal!20}2.87 & \cellcolor{teal!20}\textbf{2.53} & \cellcolor{teal!20}2.84 & \cellcolor{teal!20}2.30 & \cellcolor{teal!20}2.82 & \cellcolor{teal!20}2.32 & \cellcolor{teal!20}2.84 & \cellcolor{teal!20}2.38 \\
\bottomrule
\end{tabular}
\end{threeparttable}
\end{table*}

\paragraph{Derivation of Layer Breadth}
To formalize the derivation, we introduce an auxiliary quantity $G(k')$,
which denotes the maximum layer breadth of a subtree whose root node has
maximum allowable breadth $k'$ under the same expansion rule.
Here, the bias term $[i \neq 0]$ in Eq.~\ref{eq:breadth} is taken into account
when considering deeper layers.

Let $m' = \lfloor \log_2(k') \rfloor$.
The quantity $G(k')$ satisfies the following recursive characterization:
\begin{equation}
\label{eq:def_g}
G(k') =
\begin{cases}
\displaystyle
\sum_{i = 1}^{m'}
G\!\left(\left\lfloor \dfrac{k'}{2^i} \right\rfloor \right)
+ k' - m',
& k' > 1, \\[6pt]
1, & k' = 1.
\end{cases}
\end{equation}

Starting from the leaves and aggregating contributions bottom-up,
the maximum layer breadth of the entire Logits Tree with root breadth $b_0 = k$
can then be written as
\begin{equation}
\label{eq:def_l}
\mathcal{L}(k) =
\begin{cases}
\displaystyle
\sum_{i = 0}^{m-1}
G\!\left(\left\lfloor \dfrac{k}{2^i} \right\rfloor \right)
+ k - m,
& k > 1, \\[6pt]
1, & k = 1.
\end{cases}
\end{equation}

For brevity, we omit the intermediate algebraic steps; the closed-form
expression in Eq.~\ref{eq:max_breadth} can be obtained by resolving the
recursions in Eq.~\ref{eq:def_g} and Eq.~\ref{eq:def_l}.
It is also straightforward to see that the maximum layer breadth first occurs
at depth $\lfloor \log_2(k) \rfloor + 1$.
Beyond this depth, all node breadths reduce to 1 and the tree no longer expands,
whereas before this depth, there always exists at least one node with breadth
greater than 1, causing the layer size to continue increasing.

\paragraph{Upper Bound of Layer Breadth}
For $0 \le i \le m - 1$, we have $i + 2 \le m + 1$, which yields
$$
\begin{aligned}
\sum_{i = 0}^{m - 1}\alpha_i(i+2)2^i & \le (m+1)\sum_{i = 0}^{m - 1}\alpha_i\cdot 2^i \\
& = (m+1)\cdot r.
\end{aligned}
$$

Since $r = k - 2^{m}$, it follows that
$$
\begin{aligned}
\mathcal{L}(k) & \le (m+1)\cdot r+2^{m}\cdot m+1 \\
& = (m+1)k-2^m+1.
\end{aligned}
$$

When $k = 2^{m}$, we have $m = \lceil \log_2(k) \rceil$, and thus
$$\mathcal{L}(k)=2^m\cdot m+1=k\cdot\lceil\log_2(k)\rceil+1.$$

When $k > 2^{m}$, we have $m = \lceil \log_2(k) \rceil - 1$, which leads to
$$
\begin{aligned}
\mathcal{L}(k) & \le (m+1)k-2^m+1 \\
& = \lceil\log_2(k)\rceil\cdot k - 2^m+ 1 \\
& < k\lceil\log_2(k)\rceil + 1.
\end{aligned}
$$

This completes the proof that $\mathcal{L}(k) \le k\lceil \log_2(k) \rceil + 1$, with equality holding if and only if $k$ is a power of two.

\section{Comparison with EAGLE-3}
\label{sec:comparison_with_eagle3}

As a model-free method, RACER is orthogonal to model-based methods like EAGLE-3. While EAGLE-3 benefits from extensive training and model-based optimizations, RACER provides significant acceleration without requiring any additional training. Here, we explore in which scenarios RACER can complement EAGLE-3, potentially providing better acceleration in a hybrid setting with minimal modification to EAGLE-3.

In the supplementary results with Qwen3, we include several reasoning tasks: GSM8K~\citep{cobbe2021gsm8k}, AIME~\citep{aime_1983_2024}, and MATH~\citep{hendrycksmath2021}, using AngelSlim's re-implementation of EAGLE-3 model weights. We select 250 (+2) samples from each dataset: the first 250 samples for GSM8K, 250 random samples for AIME (from 1983 to 2024), and 50 random samples across levels 1 to 5, and 2 samples from level ? for MATH. 

Table~\ref{tab:mix_a800_eagle_reasoning} shows that EAGLE-3 achieves higher MAT across all settings, which is expected given its task-specific training and model-based predictive capability.
However, RACER attains comparable or even higher speedups in several cases (e.g., Qwen3-8B on GSM8K/AIME and Qwen3-14B across all tasks), despite having lower MAT. This highlights a key advantage of model-free decoding: RACER incurs nearly constant draft-generation cost, so its speedup does not degrade as sharply as computation-heavy model-based methods when model size increases.
Across Qwen3-14B and Qwen3-32B, RACER maintains stable acceleration (2.38-2.72$\times$), while EAGLE-3's speedup remains limited by verification overhead.
These results suggest that RACER's speculative drafting is highly efficient even for long reasoning tasks, and that its acceleration stems from computational structure rather than model training.
Most importantly, this comparison illustrates that RACER and EAGLE-3 offer orthogonal strengths:
EAGLE-3 excels when high-quality next-step predictions are available via training.
RACER excels when low overhead and robustness across tasks and model sizes are required.

\end{document}